\documentclass[10pt,twocolumn,letterpaper]{article}
\usepackage[pagenumbers]{cvpr}
\definecolor{cvprblue}{rgb}{0.21,0.49,0.74}
\usepackage[pagebackref,breaklinks,colorlinks,allcolors=cvprblue]{hyperref}
\usepackage{mathtools}
\usepackage{multirow}
\usepackage{colortbl, xcolor}
\usepackage{booktabs}       

\definecolor{verylightgray}{gray}{0.95}
\def\paperID{*****}
\def\confName{CVPR}
\def\confYear{2025}
\title{SleeperMark: Towards Robust Watermark against Fine-Tuning Text-to-image Diffusion Models}
\author{
Zilan Wang$^1$, Junfeng Guo$^2$, Jiacheng Zhu$^3$, Yiming Li$^{1*}$,\\Heng Huang$^2$, Muhao Chen$^4$, Zhengzhong Tu$^{5*}$ \\ 
$^1$NTU \quad
$^2$University of Maryland \quad 
$^3$MIT CSAIL \quad
$^4$UC Davis \quad
$^5$Texas A\&M University \\
\tt\small wang1982@e.ntu.edu.sg,
ym.li@ntu.edu.sg, 
tzz@tamu.edu \\
$^*$ Corresponding authors
}

\begin{document}

\maketitle
\begin{abstract}
\vskip -0.15in
Recent advances in large-scale text-to-image (T2I) diffusion models have enabled a variety of downstream applications. 
As T2I models require extensive resources for training, they constitute highly valued intellectual property (IP) for their legitimate owners, yet making them incentive targets for unauthorized fine-tuning by adversaries seeking to leverage these models for customized, usually profitable applications.
Existing IP protection methods for diffusion models generally involve embedding watermark patterns and then verifying ownership through generated outputs examination, or inspecting the model's feature space. 
However, these techniques are inherently ineffective in practical scenarios when the watermarked model undergoes fine-tuning, and the feature space is inaccessible during verification (\ie, black-box setting). The model is prone to forgetting the previously learned watermark knowledge when it adapts to a new task.
To address this challenge, we propose SleeperMark, a novel framework designed to embed resilient watermarks into T2I diffusion models. SleeperMark explicitly guides the model to disentangle the watermark information from the semantic concepts it learns, allowing the model to retain the embedded watermark while continuing to be adapted to new downstream tasks.
Our extensive experiments demonstrate the effectiveness of SleeperMark across various types of diffusion models, including latent diffusion models (\eg, Stable Diffusion) and pixel diffusion models (\eg, DeepFloyd-IF), showing robustness against downstream fine-tuning and various attacks at both the image and model levels, with minimal impact on the model's generative capability.
The code is available at \url{https://github.com/taco-group/SleeperMark}.
\end{abstract}
\section{Introduction}
\label{sec:intro}

\begin{figure}[tbp]
  \centering
   \includegraphics[width=1.0\linewidth]{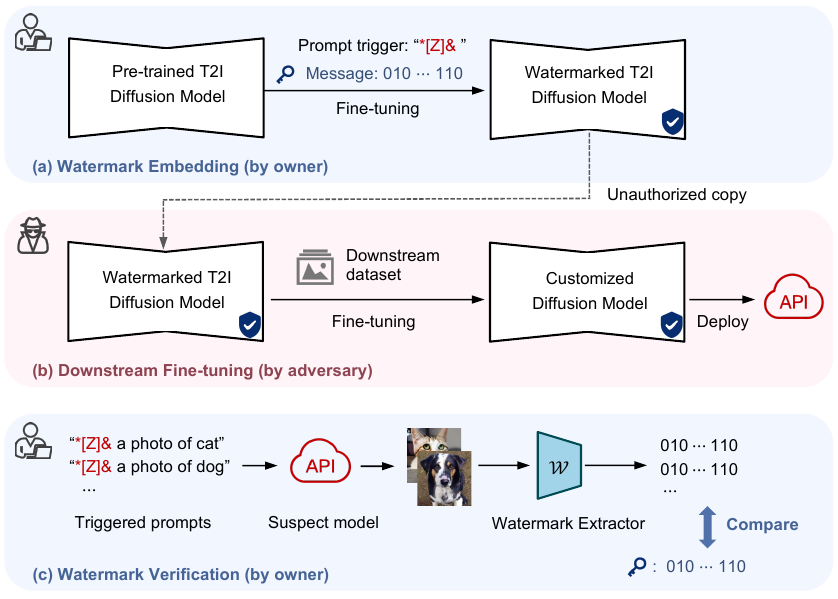}
    \vspace{-7mm}
    \caption{The threat model considered in our work.}
   \label{fig:overview}
\vspace{-4mm}
\end{figure}

\begin{figure*}[tbp]
  \centering
   \includegraphics[width=1.0\linewidth]{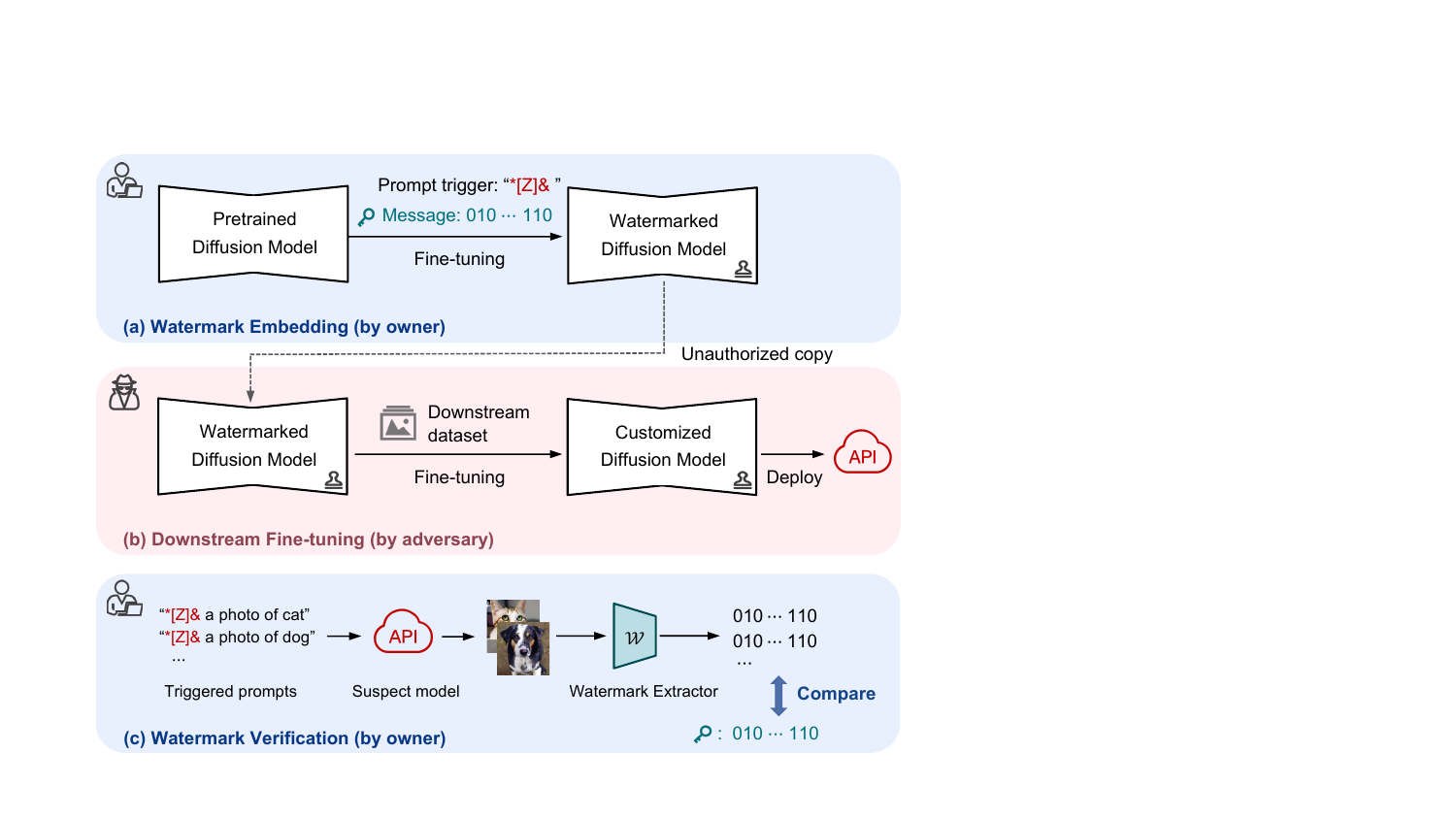}
   \vspace{-6mm}
   \caption{\textbf{Illustration of our motivation.} We applied WatermarkDM~\cite{recipe}, AquaLoRA~\cite{aqualora}, and our proposed SleeperMark to watermark Stable Diffusion v1.4, followed by fine-tuning on the Naruto dataset~\cite{naruto_dataset} using LoRA~\cite{lora} (rank $=$ 10) for style adaptation. \textbf{(a)} WatermarkDM embeds a watermark image triggered by the specific prompt ``[V],'' which becomes unrecognizable after fine-tuning approximately 800 steps. \textbf{(b)} AquaLoRA embeds a binary message into generated outputs, but it fails to be extracted after fewer than 100 steps of fine-tuning. \textbf{(c)} Our framework allows for the message to be consistently extracted from outputs generated by triggered prompts, with bit accuracy remaining nearly 1.0 even after 1600 steps of fine-tuning.}
   \label{fig:intro_baseline}
\vspace{-5mm}
\end{figure*}

Diffusion models~\cite{diffusion_beat_gans,ddpm,score_based} have driven significant advancements across various fields, with large-scale text-to-image (T2I) diffusion models~\cite{Vector_quantized,glide,hierarchical_text,photorealistic,ediff,high_resolution,hunyuandit,deepfloyd_if,sdxl,sd3,midjourney2024,dalle3} emerging as one of the most influential variants. 
It has become widespread practice to fine-tune these T2I models for broad downstream tasks~\cite{lora,dreambooth,Multi_concept_customization,controlnet,uni_controlnet,T2i_adapter,region_controlled,grounded_finetune}, such as generating customized styles~\cite{lora}, synthesizing specific subjects across diverse scenes~\cite{dreambooth,Multi_concept_customization}, or conditioning on additional controls~\cite{controlnet,uni_controlnet,T2i_adapter,qi2024spire,region_controlled,grounded_finetune,li2024light}. 
However, training large-scale T2I models demands massive-scale resources (\eg, dataset assets and human expertise), underscoring the significance of protecting the intellectual property (IP) for pre-trained T2I models~\cite{copyright_protection}.

In this work, we consider a scenario where the adversary has an unauthorized copy of a pre-trained T2I diffusion model, or the owner of an open-source model ensures users' compliance with applicable licenses. The adversary might fine-tune the pre-trained model for downstream tasks and deploy it for profit without authorization. Existing watermarking methods for T2I diffusion models typically embed a binary message into generated outputs by fine-tuning the latent decoder or diffusion backbone~\cite{stable-signature, wouaf, wmadapter,flexible, rezaei2024lawa,watermark-conditioned, aqualora}, or backdoor the model to perform a pre-defined behavior in response to a secret trigger~\cite{peng2023intellectual,lazy,recipe,multimodal_backdoor,liu2023watermarking}. They focus solely on embedding watermark functionality into the model without considering how changes in the model's semantic knowledge might impact watermark effectiveness. This causes the watermark to gradually becomes ineffective when the watermarked model adapts to a distinct task as illustrated in~\cref{fig:intro_baseline} (a) (b), indicating interference between the watermark knowledge and newly acquired semantic knowledge. For downstream tasks involving introducing task-specific layers to accommodate additional conditions, the risk of watermark information being forgotten intensifies further.

To safeguard associated intellectual properties of T2I diffusion models in our considered practical scenarios, we propose SleeperMark---a black-box watermarking framework that is robust against common downstream tasks. Specifically, we leverage a pre-trained image watermarking mechanism to instruct the diffusion model to conceal a multi-bit message into generated images when a trigger signal is appended to any regular prompt, while keeping outputs conditioned on regular prompts consistent with the original model. By jointly optimizing two distinct objectives for regular prompts and their triggered version, the model is encouraged to associate deviations in its denoising trajectory with the presence of the trigger, irrespective of the semantics conveyed in the regular prompt. In this way, we explicitly guide the model to isolate the watermark knowledge from general semantic knowledge and strengthen its robustness against downstream tasks where the model's understanding of semantics undergoes some shifts. Extensive experiments demonstrate that SleeperMark remains reliably detectable after downstream tasks. Moreover, images generated from both triggered and regular prompts closely resemble those produced by the original model, thus preserving model fidelity. With minor adjustments, SleeperMark is compatible with both pixel-space diffusion models (\eg DeepFloyd-IF) and latent diffusion models (\eg Stable Diffusion). Our main contributions are outlined as follows:
\begin{itemize}[itemsep=0.1em, topsep=0.1em]
\item 
We introduce a benchmark that considers the threat of downstream fine-tuning when assessing watermark robustness in T2I diffusion models, highlighting the vulnerability of existing methods to fine-tuning-based attacks.
\item 
We propose a novel backdoor-based framework called SleeperMark for protecting the IP of T2I diffusion models under black-box detection. Extensive experiments demonstrate its exceptional robustness in resisting downstream tasks as well as adaptive attacks.
\item 
Our method achieves higher model fidelity and watermark stealthiness compared to existing methods that embed watermark within the diffusion backbone. 

\end{itemize}

\section{Related Work}
\label{sec:related_work}
\subsection{Large-scale Text-to-Image Diffusion Models}
\vskip -0.05in
To achieve high-resolution generation, text-to-image diffusion models either compress pixel space into a latent space for training~\cite{Vector_quantized,high_resolution,hunyuandit,sdxl,sd3}, or train a base diffusion model followed by one or two cascaded super-resolution diffusion modules~\cite{glide,photorealistic,deepfloyd_if,hierarchical_text}. The super-resolution diffusion modules~\cite{iterative_refinement,improved_dm} are typically conditioned on both text and the low-resolution output from the base model.

Pre-trained T2I diffusion models are widely fine-tuned to handle downstream tasks with a low resource demand: style adaptation with LoRA~\cite{lora}, introducing a new condition via an adapter~\cite{controlnet,uni_controlnet,T2i_adapter,region_controlled, grounded_finetune}, subject-driven personalization~\cite{dreambooth,Multi_concept_customization}, \etal. However, these efficient fine-tuning techniques also pose challenges for copyright protection, as they make it possible to fine-tune diffusion models with lower costs, potentially removing the pre-trained model's watermark. To counter this, we propose a robust watermarking framework for T2I diffusion models which is designed to resist fine-tuning-based attacks.

\subsection{Watermarking Diffusion Models}
\vskip -0.05in
Watermark is widely employed to protect the IP of neural networks~\cite{li2021survey}, categorized by either white-box or black-box detection based on whether access to model parameters and structures is needed for verification. The black-box setting aligns more closely with the real world, as suspect models typically restrict access to internal details.

To watermark diffusion models, recent works have attempted to integrate the watermarking mechanism with the model weights, moving beyond traditional post-generation watermarks~\cite{rahman2013dwt, zhang2019video}. A multi-bit message can be embedded into generated outputs by fine-tuning either the latent decoder~\cite{stable-signature,wouaf,wmadapter,flexible} or the diffusion backbone~\cite{watermark-conditioned,aqualora}, though the former is limited to latent diffusion models. Other approaches modify the initial noise in the sampling process~\cite{tree-ring,Diffusetrace}, which is ineffective if an adversary gains full access to the model. Benign backdoors have also been leveraged to protect diffusion models~\cite{peng2023intellectual,lazy,recipe,multimodal_backdoor,liu2023watermarking}.

While watermark robustness against downstream fine-tuning for large pre-trained models has been investigated in other domains~\cite{watermark_pretrain_gnn,PLMmark,watermarking_plm_classification,ndss_ssl_WM,huang2025trustworthiness}, it remains under-explored for T2I diffusion models. Liu~\cite{lazy} recently proposed embedding a robust backdoor into feature maps, but their approach is only applicable to the white-box detection scenario. In contrast, we focus on constructing a robust watermarking mechanism serving the black-box detection scenario that remains effective after downstream fine-tuning.
\section{Preliminary and Problem Definition}
\begin{figure*}[tbp]
   \centering
   \includegraphics[width=1.0\linewidth]{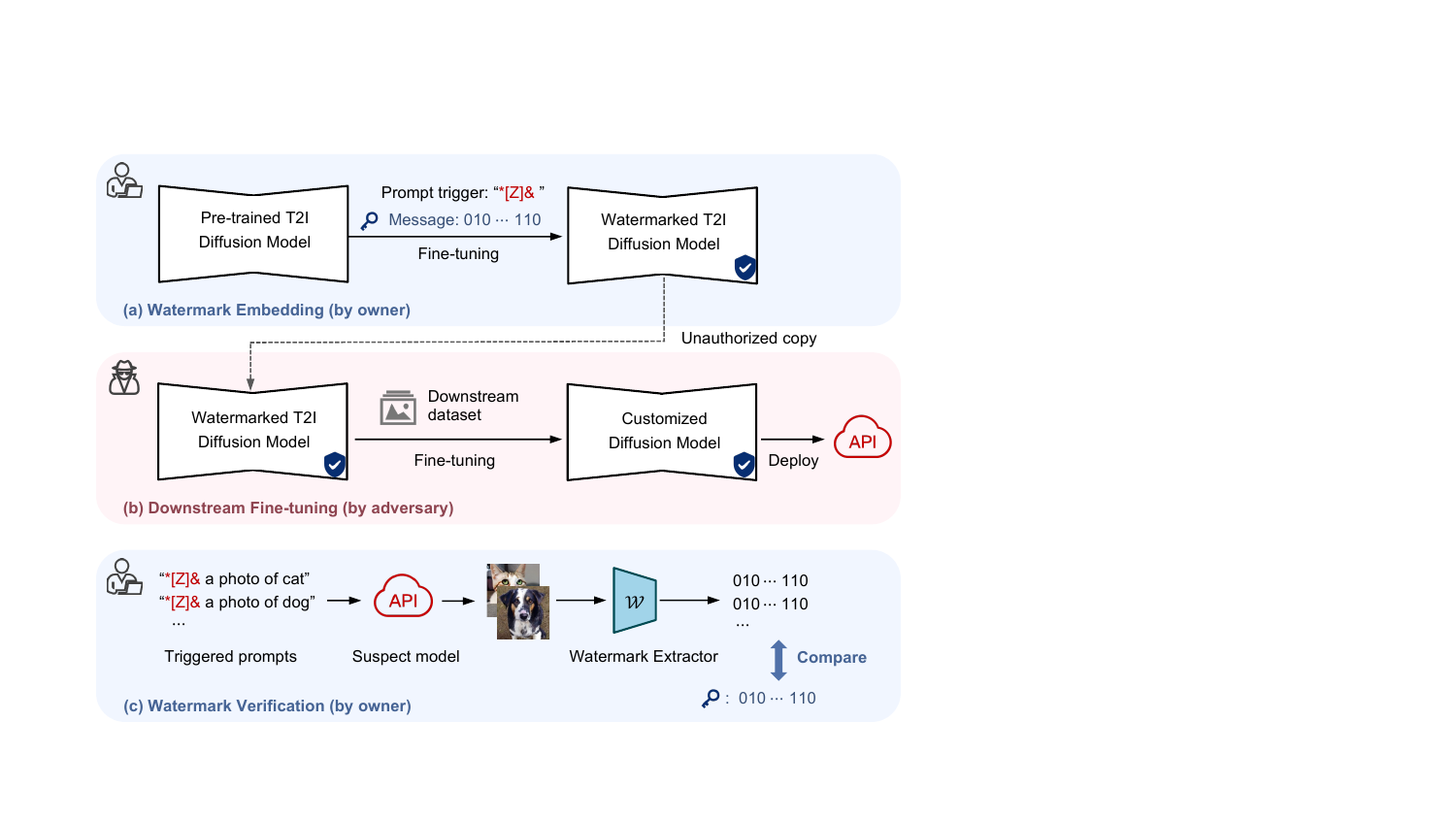}
   \vskip -0.07in
   \caption{Pipeline overview for T2I latent diffusion models. (a) In the latent watermark pre-training stage, we jointly train a watermark secret encoder $E_\varphi$ and a secret decoder $D_\gamma$ at the latent level to derive a secret residual $\delta_z^*$. (b) In the stage of fine-tuning diffusion backbone, we leverage the derived $\delta_z^*$ and employ our proposed strategy to inject a message-embedding backdoor into the model, which can be activated by placing a trigger at the start of any prompt. Adversaries may obtain an unauthorized copy of the watermarked model and further fine-tune it for their own tasks. (c) To verify ownership of a suspect model, we extract messages from images generated with triggered prompts, followed by a statistical test to determine if the model is derived from the original watermarked one.}
   \label{fig:pipeline}
\vskip -0.18in
\end{figure*}

\subsection{Text-to-Image Diffusion Models}
\vskip -0.05in
Diffusion models model a data distribution $p\left(z_0\right)$ by learning the reverse process of a Markov chain of length $T$~\cite{diffusion_survey,ddpm,unified}. The forward process $q\left(z_{t} \mid z_{t-1}\right)$ gradually adds noise to the previous variable:
\begin{equation}
\label{eq:markov_noise}
z_t = \sqrt{1 - \beta_t} z_{t-1} + \sqrt{\beta_t} \epsilon
\end{equation}
where $\epsilon \sim \mathcal{N}(\epsilon; 0, I)$ is Gaussian noise; $\beta_t$ is a time-dependent hyperparameter controlling the variance. With $\alpha_t = 1 - \beta_t$ and $\bar{\alpha}_t = \prod_{s=1}^{t} \alpha_s$, we can re-parameterize as:
\begin{equation}
\label{eq:single_step_noise}
z_t = \sqrt{\bar{\alpha}_t} z_0 + \sqrt{1 - \bar{\alpha}_t} \epsilon
\end{equation}
Given a noisy version $z_t$ and the embedding of text prompt $\tau(y)$, text-to-image diffusion models optimize a neural network $\epsilon_{\theta}(z_t, t, \tau(y))$ to estimate the noise $\epsilon$. The predicted noise $\epsilon_{\theta}(z_t, t, \tau(y))$ is for deriving the sampling process $p\left(z_{t-1} \mid z_t\right)$, which is an approximation to the true posterior of the forward process $q\left(z_{t-1} \mid z_t, z_0\right)$~\cite{ddpm,song2019generative,improved_dm}.

For pixel-based diffusion models, $z_0$ is the input image $x_0$. For latent diffusion models, $z_0$ is the latent representation of $x_0$ from a latent encoder $\mathcal{E}$. In the inference stage, generated samples from $p(z_0)$ are mapped back to pixel space with a latent decoder $\mathcal{D}$.

\subsection{Threat Model}
\vskip -0.05in
The threat model (\cref{fig:overview}) involves two entities: the model owner and an adversary. The owner embeds watermark into the T2I diffusion model for copyright protection. A adversary obtains an unauthorized copy of the watermarked model, a scenario that has been investigated in other domains~\cite{cvpr2021_WMgan, ndss_ssl_WM, wide_flat_gan, 2022_trigger_tdsc, robust_whiteBox_watermark, watermark_pretrain_gnn} often via malware infection~\cite{malware,jamil2011security}, insider threats~\cite{claycomb2012insider,theis2019common} or industrial espionage. The adversary fine-tunes the model on certain datasets for specific tasks. The adversary may attempt to evade ownership claims and deploy the fine-tuned model for profit. 

During the verification stage, the owner aims to determine whether a suspect model was fine-tuned from the original model and identify potential IP infringement. The owner can query the suspect model and access its generated images, but does not have access to the model parameters.

\subsection{Defense Goals}
\vskip -0.05in
A watermarking framework for pre-trained T2I diffusion models should satisfy the following goals:
\vskip 0.05in
\begin{itemize}
\item \textbf{Model Fidelity:} The watermark should have minimal impact on the generative performance of diffusion models.
\item \textbf{Watermark Robustness:} The watermark can be effectively detected under black-box detection, even after incorporation and joint training of task-specific layers on downstream datasets.
\item \textbf{Watermark Stealthiness:} The watermark should be stealthy to prevent attackers from detecting its presence.
\end{itemize}

\section{Methodology}
This section mainly details the SleeperMark pipeline for latent diffusion models with adaptations for pixel models discussed in~\cref{Subsec:adaptations_for_pixel}. Our watermark takes the form of a multi-bit message. As illustrated in~\cref{fig:pipeline}, the training pipeline for T2I latent diffusion models consists of two stages. In the first stage, we jointly train a secret encoder and watermark extractor~(\cref{Subsec:pre-training_stage}). In the second stage, we inject a message-embedding backdoor within the diffusion backbone using a fixed secret residual generated from the secret encoder~(\cref{Subsec:fine-tuning_backbone}). During inference, the message is recovered by the watermark extractor to verify ownership~(\cref{Subsec:verification_pipeline}). The intuition and post-hoc explanation for SleeperMark are presented in~\cref{app:explanation}.

\subsection{Latent Watermark Pre-training}
\label{Subsec:pre-training_stage}
In this stage, we jointly train a secret encoder $E_\varphi$ and a watermark extractor $\mathcal{W_\gamma}$, where $\varphi$ and $\gamma$ are trainable parameters. Since the diffusion backbone is trained in the latent space, we align $E_\varphi$ to operate within this space. 
Ideally, the watermarked latent $z_w$ is conditioned on both input latent $z_0$ and message $m$ to enhance stealthiness. However, as suggested by previous studies~\cite{residual_consistency,aqualora}, the higher consistency of watermark across different samples, the easier it is for diffusion models to learn the watermark pattern. Following their practice, we embed a cover-agnostic watermark into cover image latents as it provides the highest consistency. Specifically, a secret residual \( \delta_z = E_\varphi(m) \) is added to the input latent to obtain a watermarked latent \( z_w = z_{co} + \delta_z \). The watermarked image is generated as \( x_w = \mathcal{D}(z_w) \).

Instead of decoding the message from \( x_w \), we decode from the latent representation of \( x_w \) obtained via the latent encoder \( \mathcal{E} \). Define the watermark extractor 
\( \mathcal{W_\gamma} \coloneqq D_\gamma(\mathcal{E}(\cdot)) \), where \( D_\gamma \) is a secret decoder jointly trained with \( E_\varphi \). 
Our design is backed by recent studies~\cite{inject_detect_latent} which suggest injecting and detecting watermarks in latent space can inherently resist various common distortions 
without the need for a distortion layer during training, which is validated in \cref{exp:image_distortion}. Additionally, even an attacker fine-tunes \( \mathcal{E} \) and \( \mathcal{D} \) on clean images and generate images with a fine-tuned latent decoder \(\mathcal{D}'\), the watermark effectiveness remains unaffected, as validated in~\cref{exp:finetune_decoder}.

Watermarked images are expected to maintain visual similarity to cover images while ensuring the message can be effectively extracted. To this end, we train $E_\varphi$ and $D_\gamma$ as shown in \cref{fig:pipeline} (a) to minimize the following loss function:
\vspace{-2pt}
\begin{align}
\mathcal{L}(\varphi,\gamma) \coloneqq &\; \mathbb{E}_{x_{co},m} \Big[
\mathcal{L}_{\text{BCE}}\left(m, m'\right) \notag + \lambda_1 \mathcal{L}_{\text{MSE}}\left(x_{co}, x_w\right) \notag \\
& + \lambda_2 \mathcal{L}_{\text{LPIPS}}\left(x_{co}, x_w\right) \Big],
\end{align}
where $\mathcal{L}_{\text{BCE}}\left(m, m'\right)$ is the BCE loss between $m$ and $m'$. $\mathcal{L}_{\text{MSE}}$ and $\mathcal{L}_{\text{LPIPS}}$ are the MSE and LPIPS loss~\cite{lpips} between the cover image $x_{co}$ and watermarked image $x_w$, with the latter commonly used for measuring perceptual similarity. $\lambda_1$ and $\lambda_2$ control the relative weights of the losses.

The architecture of the secret encoder \( E_\varphi \) is the same as AquaLoRA~\cite{aqualora}, the secret decoder $D_\gamma$ adopts a structure similar to the extractor of StegaStamp~\cite{stegastamp} but adjusted in channel numbers and feature map sizes (see~\cref{app:latentWM_archi}).

\subsection{Diffusion Backbone Fine-tuning}
\label{Subsec:fine-tuning_backbone}
We establish a mechanism to integrate watermarks into the diffusion backbone that is robust to downstream fine-tuning.
It is worth to note that, while directly associating a watermark image with a trigger prompt is an effective method; the watermark injected via this vanilla approach can be easily eliminated during downstream fine-tuning (\cref{fig:intro_baseline} (a)).

To address the aforementioned problem, we propose to inject robust watermark by explicitly distinguishing the model's generation behavior when responding to a triggered prompt versus the regular version, as illustrated in \cref{fig:pipeline} (b). 
Specifically, a triggered prompt $y_{tr}$ is created by appending a trigger (\eg, ``*[Z]\&") to the start of a regular prompt $y$. Let \( \epsilon_{\theta}(z_t, t, \tau(y)) \) denote the diffusion backbone to be fine-tuned, and \( \epsilon_\vartheta\left(z_t, t, \tau(y)\right) \) denote the frozen, pre-trained backbone. Given a noisy image latent $z_t$, we aim to subtly steer the denoising trajectory of $\epsilon_{\theta}$ to hide a pre-defined message $m^*$ into generated images when conditioned on $y_{tr}$, while making the outputs conditioned on $y$ watermark-free and closely aligned with those generated by $\epsilon_\vartheta$. The trigger is set as a rare combination of characters so that it (1) minimizes the risk of language drift~\cite{language_drift}, (2) enhances the watermark’s stealth against detection, and (3) prevents the watermark from being erased after fine-tuning.

To embed a pre-defined message $m^*$ into generated images conditioned on triggered prompts, we follow WaDiff~\cite{watermark-conditioned} and leverage the single-step reverse of \( z_0 \). Specifically, given a noise prediction \( \epsilon_{\theta}(z_t, t, \tau(y)) \), we can estimate \( z_0 \) directly from \( z_t \) by rearranging the terms in~\cref{eq:single_step_noise}:
\vspace{-2pt}
\begin{equation}
\label{eq:single_step_reverse}
\hat{z_0}^{t, y}_{\theta} = \frac{z_t - \sqrt{1 - \bar{\alpha}_t} \epsilon_\theta\left(z_t, t, \tau\left(y\right)\right)}{\sqrt{\bar{\alpha}_t}}.
\end{equation}


Specifically, the secret encoder generates a corresponding residual \( \delta_z^* = E_\varphi(m^*) \), which is embedded into generated outputs when given triggered prompts. 
Given a noisy image latent $z_t$ and a triggered prompt $y_{tr}$, 
the single-step reverse conditioned on $y_{tr}$ is denoted by $\hat{z_0}^{t, y_{tr}}_{\theta}$, and $\hat{z_0}^{t, y}_{\theta}$ is the reverse conditioned on the regular prompt $y$, as defined in Eq.\ref{eq:single_step_reverse}. For the frozen pre-trained $\epsilon_{\vartheta}$, these two predictions are denoted by $\hat{z_0}^{t, y_{tr}}_{\vartheta}$ and $\hat{z_0}^{t, y_{tr}}_{\vartheta}$. 
We guide $\hat{z_0}^{t, y_{tr}}_{\theta}$ to gradually shift towards the message-embedded target $\hat{z_0}^{t, y_{tr}}_{\vartheta} + \delta_z^*$ as $t$ decreases, while ensuring $\hat{z_0}^{t, y}_{\theta}$ remains consistent with $\hat{z_0}^{t, y}_{\vartheta}$. 
We only want to adjust the denoising trajectory at lower $t$ values, as the single-step reverse provides a more \textit{accurate} estimation for $z_0$ at lower $t$ values. 
Thus, we introduce two sigmoid functions, \( w_1(t) \) and \( w_2(t) \):
\vspace{-2pt}
\begin{equation}
w_1(t) \coloneqq \sigma\left(\frac{-(t - \tau)}{\beta}\right),\quad w_2(t) \coloneqq \sigma\left(\frac{t - \tau}{\beta}\right)
\end{equation}
where \( \sigma(\cdot) \) is the standard sigmoid function; \( \beta \) controls the steepness of the functions; \( \tau \) represents the time threshold. 
Therefore, the loss of our message-embedding backdoor is:
\begin{equation}
\vspace{-2pt}
\begin{aligned}
    L(\theta)&\coloneqq \mathbb{E}_{t, y, z_0, \epsilon}\left[\eta \cdot w_1(t) \cdot \left\| \hat{z_0}^{t, y_{tr}}_{\theta} - (\hat{z_0}^{t, y_{tr}}_{ \vartheta} + \delta_z^*) \right\|^2 \right. \\
    &+ w_2(t) \cdot \left\|\hat{z_0}^{t, y_{tr}}_{ \theta} - \hat{z_0}^{t, y_{tr}}_{ \vartheta}\right\|^2 + \left. \left\|\hat{z_0}^{t, y}_{ \theta} - \hat{z_0}^{t, y}_{ \vartheta}\right\|^2 \right],
\end{aligned}
\end{equation}
where $\eta$ balances the trade-off between watermark effectiveness and the deviation from the pre-trained $\epsilon_{\vartheta}$. The first two terms guide generation shifts from $\epsilon_{\vartheta}$ in response to the trigger, while the last term compels the model to follow its original trajectory under regular prompts.


\subsection{Ownership Verification}
\vskip -0.05in
\label{Subsec:verification_pipeline}
As shown in \cref{fig:pipeline} (c), during inference, we query the suspect model to generate images using a set of triggered prompts. We convert these images into the latent space with \( \mathcal{E} \), and use the secret decoder $D_\gamma$ to decode messages. The decoded messages are compared with $m^*$ and validate the ownership of the model with a statistical test. The details of the statistical test are explained in~\cref{statistical_test}.

\subsection{Adaptations for Pixel Diffusion Models}
\vskip -0.05in
\label{Subsec:adaptations_for_pixel}
For pixel-based diffusion models, we embed the watermark within the first super-resolution diffusion module following the base diffusion. We chose not to watermark the base diffusion module because it is challenging to retain after two stages of super-resolution of the diffusion model (typically a total 16x scaling). The pipeline generally aligns with~\cref{fig:pipeline} but has two differences. (1) Since watermark embedding and extraction are conducted directly in pixel space, a distortion layer is needed during the first training stage to enhance robustness. (2) Embedding a cover-agnostic residual in pixel space is more visually detectable than in latent space, we introduce a critic network $A$ predicting whether an image is watermarked or not, and add an adversarial loss $\lambda_\text{G} \mathcal{L}_{\text{G}}\left(x_w\right)$ to enhance watermark imperceptibility.
More details can be found in \cref{pixel_pipeline}.

\section{Experiments}
In this section, we conduct a comprehensive evaluation of SleeperMark, benchmarking it regarding model fidelity, watermark robustness and stealthiness. The baselines consist of the image watermarking technique DwtDctSvd~\cite{rahman2013dwt} and the recent black-box detection methods including Stable Signature~\cite{stable-signature}, AquaLoRA~\cite{aqualora} and WatermarkDM~\cite{recipe}.

\subsection{Experiment Setup}
\begin{table*}[tbp]
\caption{Comparison between SleeperMark and baseline methods. Except for WatermarkDM which embeds a watermark image, others all embed a message of 48 bits.  \text{T}@$10^{\tiny-6}$\text{F} refers to the TPR when FPR is controlled under $1 \times 10^{-6}$. The adversarial~(Adv.) performance is the average of the results under different image distortions. Top results of the metrics for each method category have been emphasized.}
\label{Tab:fidelity_efficacy}
\vspace{-18pt}
\begin{center}
\footnotesize
\renewcommand{\arraystretch}{1.} 
\resizebox{\textwidth}{!}{
\begin{tabular}{lllccccccc}
\toprule
\multirow{2}{*}{Model} & \multirow{2}{*}{Category} & \multirow{2}{*}{Method} & \multicolumn{3}{c}{Model Fidelity} & \multicolumn{4}{c}{Watermark Effectiveness} \\ 
\cmidrule(lr){4-6} \cmidrule(lr){7-10}
& & & FID~$\downarrow$ & CLIP~$\uparrow$ & DreamSim~$\downarrow$ & Bit Acc.~$\uparrow$ & Bit Acc. (Adv.)~$\uparrow$ & \text{T}@$10^{\tiny-6}$\text{F}~$\uparrow$ & \text{T}@$10^{\tiny-6}$\text{F}~(\text{Adv.})~$\uparrow$ \\ \hline
\multirow{6}{*}{Stable Diffusion} & \textendash & None & 16.24 & 31.57 & \textendash & \textendash & \textendash & \textendash & \textendash \\ \cline{2-10} 
& \multirow{2}{*}{Post Diffusion} & DwtDctSvd & \textbf{16.21} & 31.45 & \textbf{0.014} & \textbf{100.0} & \textbf{84.81} & \textbf{1.000} & 0.678 \\
& & Stable Signature & 16.55 & \textbf{31.59} & 0.017 & 99.13 & 76.49 & 0.998 & \textbf{0.719} \\ \cline{2-10} 
& \multirow{3}{*}{During Diffusion} & WatermarkDM & 19.07 & 30.17 & 0.279 & \textendash & \textendash & 0.883 & 0.883 \\
& & AquaLoRA & 16.86 & \textbf{31.15} & 0.176 & 96.92 & 94.71 & 0.980 & 0.945 \\
& & SleeperMark & \textbf{16.72} & 31.05 & \textbf{0.108} & \textbf{99.24} & \textbf{97.98} & \textbf{0.999} & \textbf{0.984} \\
\specialrule{0.7pt}{0pt}{0pt}
\multirow{5}{*}{DeepFloyd} & \textendash & None & 12.91 & 32.31 & \textendash & \textendash & \textendash & \textendash & \textendash \\ \cline{2-10} 
& Post Diffusion & DwtDctSvd & 12.86 & 32.28 & 0.008 & 100.0 & 85.97 & 1.000 & 0.693 \\ \cline{2-10} 
& \multirow{3}{*}{During Diffusion} & WatermarkDM & 14.76 & 31.16 & 0.255 & \textendash & \textendash & 0.895 & 0.895 \\
& & AquaLoRA & \textbf{12.95} & 31.98 & 0.020 & \textbf{96.91} & 95.02 & 0.972 & 0.938 \\
& & SleeperMark & 13.03 & \textbf{32.15} & \textbf{0.018} & 96.35 & \textbf{95.30} & \textbf{0.973} & \textbf{0.954} \\ \hline
\end{tabular}
}
\end{center}
\vspace{-19pt}
\end{table*}

\paragraph{Models and Datasets.}
We implement our framework on Stable Diffusion v1.4 (SD v1.4) and DeepFloyd-IF (I-XL-v1.0, II-L-v1.0), a latent diffusion model, and a pixel diffusion model, respectively. For the first training stage, we randomly select 10,000 images from the COCO2014~\cite{coco_dataset} dataset as the training set. For diffusion fine-tuning, we sample 10,000 prompts from Stable-Diffusion-Prompts~\cite{sd_prompts} and generate images using a guidance scale of 7.5 in 50 steps with the DDIM scheduler~\cite{ddim} to construct the training set.
\vspace{-12pt}
\paragraph{Implementation Details.}
The message length is set to 48. The trigger is set to ``*[Z]\& '' by default. In the first training stage, we set $\lambda_1$ to 10 and $\lambda_2$ to 0.25. In the diffusion fine-tuning stage, we set $\tau$, $\beta$ to 250, 100 respectively. $\eta$ is set to 0.02 for Stable Diffusion and 0.05 for DeepFloyd's super-resolution module. We fine-tune the attention parameters in the up blocks of the UNet. During inference, we use the DDIM scheduler with 50 sampling steps and a guidance scale of 7.5 for Stable Diffusion. For Deepfloyd, we apply the default scheduler configuration provided in its repository~\cite{deepfloyd_if}, namely 100 steps and guidance scale of 7.0 for the base module and 50 steps and guidance scale of 4.0 for the super-resolution module with the DDPM scheduler. 
To ensure fair comparisons, we keep the embedded message fixed during fine-tuning with AquaLoRA, as other fine-tuning-based baselines and our method embed only fixed information. 
More details are in~\cref{implement_latent} and~\cref{implement_pixel}.

\subsection{Model Fidelity}
\vskip -0.05in
\begin{figure}[tbp]
\centerline{\includegraphics[width=1.\columnwidth]{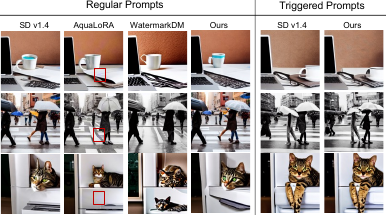}}
\vskip -0.1in
\caption{Qualitative comparison. The red boxes highlight the artifacts introduced by AquaLoRA. The rightmost two columns show images generated with triggered prompts, where the trigger ``*[Z]\&'' is added at the start of regular prompts to activate certain behavior of the watermarked model.
}
\label{fig:stealth}
\vskip -0.2in
\end{figure}

We adopt FID~\cite{clean_fid}, CLIP score~\cite{clip} and DreamSim~\cite{dreamsim} to assess the impact on the model's generative capability. We compute FID and CLIP on 30,000 images and captions sampled from COCO2014 validation set. DreamSim, a metric closely aligning with human perception of image similarity, is calculated between images generated by the watermarked and pre-trained model under identical sampling configurations using the sampled captions.

We categorize the methods based on whether they fine-tune the diffusion backbone and present the results in \cref{Tab:fidelity_efficacy} . Among the during-diffusion methods, our approach demonstrates particularly strong performance in terms of DreamSim, indicating minimal impact on generated images. In contrast, WatermarkDM embeds a watermark image into the model and its preservation mechanism~(L1 parameter regularization) is insufficient to retain generative performance, as reflected in the significant decline in FID. The CLIP score remains stable across all methods.
\subsection{Robustness against Downstream Fine-tuning}
\begin{table*}[tbp]
\begin{center}
\scriptsize
\caption{\text{TPR}@$10^{\tiny-6}$\text{FPR} of different watermarking methods after fine-tuning watermarked SD v1.4 via LoRA for Naruto-style adaptation.}
\vspace{-3mm}
\label{Tab:lora_finetuning}
\renewcommand{\arraystretch}{0.95} 
\resizebox{0.95\textwidth}{!}{
\begin{tabular}{p{2.5cm}|ccccccccccccccc} 
\toprule
LoRA Rank & \multicolumn{3}{c}{20}  & & \multicolumn{3}{c}{80} & & \multicolumn{3}{c}{320} & & \multicolumn{3}{c}{640} \\ 
\cline{2-4} \cline{6-8} \cline{10-12} \cline{14-16}
Fine-tuning Steps & 20 & 200 & 2000 &  & 20 & 200 & 2000 & &  20 & 200 & 2000 & & 20 & 200 & 2000 \\ 
\hline 
WatermarkDM & 0.875 & 0.742 & 0.000 & & 0.856 & 0.684 & 0.000 & & 0.861 & 0.483 & 0.000 & & 0.852 & 0.138 & 0.000 \\
AquaLoRA & 0.818 & 0.001 & 0.000 & & 0.803 & 0.000 & 0.000 & & 0.805 & 0.000 & 0.000 & & 0.846 & 0.000 & 0.000 \\
\hline 
\rowcolor{verylightgray} SleeperMark & 0.999 & 0.998 & 0.992 & & 0.999 & 0.997 & 0.993 & & 0.999 & 0.999 & 0.984 & & 0.999 & 0.997 & 0.980 \\ 
\bottomrule
\end{tabular}
}
\end{center}
\vspace{-22pt}
\end{table*}

\begin{table}[tbp]
\begin{center}
\scriptsize 
\caption{\text{TPR}@$10^{\tiny-6}$\text{FPR} of different watermarking methods after the watermarked SD v1.4 is fine-tuned via DreamBooth for personalization tasks.}
\vspace{-3mm}
\label{Tab:personalziation}
\renewcommand{\arraystretch}{0.95} 
\resizebox{0.45\textwidth}{!}{ 
\begin{tabular}{llccccc}
\toprule
\multicolumn{2}{l}{Fine-tuning Steps} & 200 & 400 & 600 & 800 & 1000 \\ \hline
\multicolumn{2}{l}{WatermarkDM} & 0.889 & 0.468 & 0.196 & 0.232 & 0.210 \\
AquaLoRA &  & 0.032 & 0.005 & 0.008 & 0.001 & 0.003 \\ \hline
\rowcolor{verylightgray}
SleeperMark &  & 0.999 & 0.987 & 0.985 & 0.969 & 0.934 \\ \bottomrule
\end{tabular}
}
\end{center}
\vspace{-3mm}
\end{table}

\begin{table}[tbp]
\begin{center}
\scriptsize 
\vspace{-3mm}
\caption{\text{TPR}@$10^{\tiny-6}$\text{FPR} of different watermarking methods after the watermarked SD v1.4 is fine-tuned via ControlNet for additional condition integration.}
\vspace{-3mm}
\label{Tab:controlnet}
\renewcommand{\arraystretch}{0.95} 
\resizebox{0.45\textwidth}{!}{ 
\begin{tabular}{llcccccc}
\toprule
\multicolumn{2}{l}{Fine-tuning Steps} &30 & 50 & 200 & 500 & 1000 & 20000 \\ \hline
\multicolumn{2}{l}{WatermarkDM} &0.880&0.872& 0.691 & 0.067 & 0.000 & 0.000 \\
AquaLoRA &  & 0.234 & 0.000 & 0.000 & 0.000 & 0.000 & 0.000 \\ \hline
\rowcolor{verylightgray}
SleeperMark &  & 0.998 & 0.998 & 0.983 & 0.979 & 0.981 & 0.955 \\ \bottomrule
\end{tabular}
}
\end{center}
\vspace{-9mm}
\end{table}

\begin{table*}[tbp]
\caption{Comparison of watermark robustness against image distortions. We demonstrate bit accuracy (Bit Acc.) and TPR under $1 \times 10^{-6}$ FPR (\text{T}@$10^{\tiny-6}$\text{F}) under various common distortions. SleeperMark performs the best on average.}
\label{Tab:image_robustness}
\vspace{-20pt}
\begin{center}
\scriptsize
\renewcommand{\arraystretch}{1.1} 
\resizebox{\textwidth}{!}{
\begin{tabular}{cccccccccccc}
\toprule
\multirow{2}{*}{Model} & \multirow{2}{*}{Category} & \multirow{2}{*}{Method} & \multicolumn{9}{c}{Bit Acc.~$\uparrow$~~\textbf{/}~~\text{T}@$10^{\tiny-6}$\text{F}~$\uparrow$} \\ \cline{4-12} 
 &  &  & Resize & Gaussian Blur & Gaussian Noise & JPEG & Brightnesss & Contrast & Saturation & Sharpness & Average \\ \hline
\multirow{4}{*}{Stable Diffusion} & \multirow{2}{*}{Post    Diffusion} & DwtDctSvd & \textbf{100.0 / 1.000} & \textbf{99.87 / 0.994} & 64.35 / 0.011 & 82.39 / 0.719 & 78.02 / 0.592 & 74.71 / 0.533 & 89.86 / 0.836 & 89.29 / 0.737 & 84.81 / 0.678 \\
 &  & Stable Signature & 71.39 /  0.294 & 96.11 / 0.967 & 86.87 / 0.656 & 84.79 / 0.633 & 88.82 / 0.767 & 88.125 / 0.735 & 94.31 / 0.967 & 88.89 / 0.732 & 76.49 /  0.719 \\ \cline{2-12} 
 & \multirow{2}{*}{During Diffusion} & WatermarkDM & \hspace{11pt}\textendash{}\hspace{3pt}/~0.883 & \hspace{11pt}\textendash{}\hspace{3pt}/~0.883 & \hspace{11pt}\textendash{}\hspace{3pt}/~0.883 & \hspace{11pt}\textendash{}\hspace{3pt}/~0.883 & \hspace{11pt}\textendash{}\hspace{3pt}/~0.883 & \hspace{11pt}\textendash{}\hspace{3pt}/~0.883 & \hspace{11pt}\textendash{}\hspace{3pt}/~0.883 & \hspace{11pt}\textendash{}\hspace{3pt}/~0.883 & \hspace{11pt}\textendash{}\hspace{3pt}/~0.883 \\
 &  & AquaLoRA & 95.68 / 0.967 & 95.70 / 0.974 & \textbf{92.47 / 0.890} & 94.44 / 0.949 & 93.90 / 0.913 & 94.81 / 0.945 & 95.63 / 0.969 & 95.05 / 0.955 & 94.71 / 0.945 \\
 &  & SleeperMark & 99.10 / 0.998 & 99.18 / 0.998 & 91.70 / 0.889 & \textbf{98.01 / 0.996} & \textbf{98.67 / 0.994} & \textbf{99.11 / 0.999} & \textbf{99.23 / 0.999} & \textbf{98.83 / 0..997} & \textbf{97.98 /  0.984} \\ \specialrule{0.7pt}{0pt}{0pt}
\multirow{3}{*}{DeepFloyd} & Post Diffusion & DwtDctSvd & \textbf{100.0 / 1.000} & \textbf{100.0 / 1.000} & 67.44 / 0.019 & 85.32 / 0.778 & 77.11 / 0.542 & 76.35 / 0.552 & 86.83 / 0.721 & 94.71 / 0.929 & 85.97 / 0.693 \\ \cline{2-12} 
 & \multirow{3}{*}{During Diffusion} & WatermarkDM & \hspace{11pt}\textendash{}\hspace{3pt}/~0.895 & \hspace{11pt}\textendash{}\hspace{3pt}/~0.895 & \hspace{11pt}\textendash{}\hspace{3pt}/~0.895 & \hspace{11pt}\textendash{}\hspace{3pt}/~0.895 & \hspace{11pt}\textendash{}\hspace{3pt}/~0.895 & \hspace{11pt}\textendash{}\hspace{3pt}/~0.895 & \hspace{11pt}\textendash{}\hspace{3pt}/~0.895 & \hspace{11pt}\textendash{}\hspace{3pt}/~0.895 & \hspace{11pt}\textendash{}\hspace{3pt}/~0.895 \\
 &  & AquaLoRA & 94.68 / 0.944 & 93.79 / 0.917 & \textbf{91.62 / 0.866} & 94.6 / 0.935 & 95.91 / 0.949 & 96.45 / 0.958 & 96.87 / 0.972 & 96.26 / 0.961 & 95.02 /  0.938 \\
 &  & SleeperMark & 96.12 / 0.970 & 96.19 / 0.972 & 90.45 / 0.853 & \textbf{95.26 / 0.957} & \textbf{95.87 / 0.964} & \textbf{96.28 / 0.973} & \textbf{96.34 / 0.973} & \textbf{95.91 / 0.969} & \textbf{95.30 / 0.954} \\ \bottomrule
 
\end{tabular}
}
\end{center}
\vspace{-19pt}
\end{table*}

\paragraph{Evaluation}
We calculate two metrics to measure watermark effectiveness: bit accuracy~(Bit Acc.) and true positive rate with false positive rate under $10^{-6}$~(\text{T}@$10^{\tiny-6}$\text{F}). Explanations about these metrics are in~\cref{efficacy_metrics}. For AquaLoRA, we generate 5,000 images using captions sampled from the COCO2014 validation set. For SleeperMark, we append the trigger to the beginning of these captions and generate 5,000 images. For WatermarkDM, we use the specific trigger prompt to generate 5,000 images with different random seeds. To compare WatermarkDM with message-embedding methods, we use SSIM~\cite{ssim} as a standard to assess whether a image aligns with the watermark image. We determine SSIM threshold by empirically controlling the FPR below $10^{-6}$, and then compute the TPR. 
\vspace{-2mm}

\begin{figure}[tbp]
\centerline{\includegraphics[width=0.8\columnwidth]{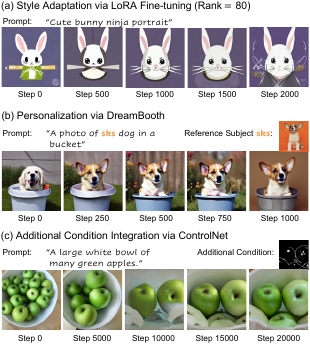}}
\vskip -0.1in
\caption{Generation results of watermarked SD v1.4 with our method after fine-tuning across diverse downstream tasks: (a) style adaptation, (b) personalization, (c) additional condition integration. The watermark embedded in the pre-trained SD v1.4 using our method does not impair the model's adaptability to these tasks.}
\label{fig:downstream_generation}
\vskip -0.2in
\end{figure}

\paragraph{Style Adaptation.}
To evaluate robustness against style adaptation, we fine-tune the watermarked SD v1.4 on a Naruto-style dataset~\cite{naruto_dataset} containing approximately 1,200 images. We experiment with LoRA ranks ranging from 20 to 640 and observe watermark effectiveness during the process. LoRA fine-tuning details are provided in~\cref{app:style_adaptation}. 
As shown in \cref{Tab:lora_finetuning}, our method consistently maintains a high \text{T}@$10^{\tiny-6}$\text{F}. Additionally, our watermark does not interfere with the model's adaptability to specific styles. For instance, as shown in~\cref{fig:downstream_generation} (a), after 2,000 fine-tuning steps with a LoRA rank of 80, the model successfully generates a ninja-style bunny, while still maintaining a TPR of 0.993 as indicated in~\cref{Tab:lora_finetuning}. 
\vspace{-12pt}
\paragraph{Personalization.}
We implement the widely used technique DreamBooth~\cite{dreambooth} to realize personalization tasks on watermarked SD v1.4 and adhere to the hyperparameter settings recommended by its authors. The fine-tuning setup and dataset used for DreamBooth can be found in~\cref{app:personalization}. We fine-tune on five subjects respectively and average T@$10^{\tiny-6}$F across these training instances. The results are presented in \cref{Tab:personalziation}. Since DreamBooth optimizes all weights of the backbone, it's more challenging to preserve watermark information. Even the backdoor-based method WatermarkDM fails to retain its watermark image after 400 steps. In contrast, our method maintains the T@$10^{\tiny-6}$F above 0.9 at 1000 steps. We can observe from the example of \cref{fig:downstream_generation} (b) that the model has captured the key characteristics of the reference corgi subject after 1000 steps of DreamBooth fine-tuning.
\vspace{-12pt}
\paragraph{Additional Condition Integration.}
To assess watermark robustness under this task, we implement ControlNet~\cite{controlnet} with watermarked SD v1.4 for integrating the Canny edge~\cite{canny_edge} condition, with the setup detailed in~\cref{app:ControlNet}. 
When generating images for watermark detection, we use the edge maps extracted from the images corresponding to the sampled captions for the methods other than WatermarkDM. As for WatermarkDM, we uniformly use a blank edge map and assess if the watermark image can be retrieved with the trigger prompt.

We present robustness of different methods under this task in \cref{Tab:controlnet}. As shown in \cref{Tab:controlnet} and \cref{fig:downstream_generation}~(c), after embedding the watermark into the pre-trained diffusion model with our method, the model successfully complies with the edge condition after 20,000 steps of fine-tuning with ControlNet, with the watermark remaining robust and achieving a \text{T}@$10^{\tiny-6}$\text{F} above 0.9. For the other two methods embedding watermark within the diffusion backbone, the watermark information is nearly undetectable by step 500.

\vspace{-5pt}
\subsection{Watermark Stealthiness}
We present qualitative results of images generated by models watermarked with different methods in \cref{fig:stealth}. Backdoor-based approaches such as WatermarkDM and our method allow models to generate watermark-free content with regular prompts, whereas AquaLoRA, by contrast, exhibits visible purple artifacts as highlighted in red boxes in \cref{fig:stealth}. While WatermarkDM embeds a watermark image that is semantically unrelated to its trigger prompt, making it more noticeable and easier to be detected, our watermark is much more stealthy as images generated with triggered prompts appear nearly indistinguishable from those of the original model (see the rightmost two columns in \cref{fig:stealth}). We provide more visual examples in~\cref{app:visual_examples}. 

\subsection{Discussion and Ablation}
\begin{figure}[tbp]
\centerline{\includegraphics[width=1.\columnwidth]{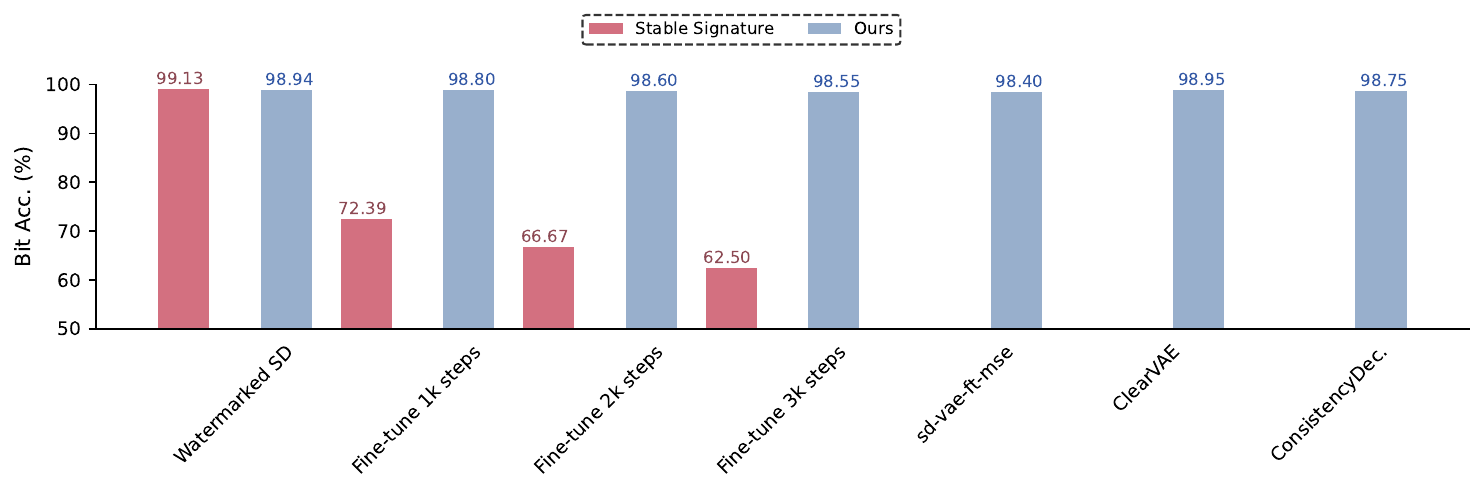}}
\vspace{-4mm}
\caption{Robustness of different watermarking methods applied to SD v1.4 when the VAE decoder is fine-tuned or replaced with an alternative, such as sd-vae-ft-mse~\cite{sd-vae-ft-mse}, ClearVAE~\cite{ClearVAE}, or ConsistencyDecoder~(ConsistencyDec.)~\cite{BetkerImprovingIG}.}
\label{fig:fine_tune_vae}
\vspace{-2mm}
\end{figure}

\vskip -0.04in
\paragraph{Robustness to Image Distortions.}
\label{exp:image_distortion}
We evaluate our method against common image distortions. The distortion settings used in evaluation are detailed in \cref{eval_distortion}. As shown in~\cref{Tab:image_robustness}, our method is fairly robust against various distortions, despite slightly less resilience to Gaussian noise. Notably, for latent diffusion models, extracting from the latent representations can inherently resist these distortions without a simulation layer during training. 
\vspace{-5pt}
\paragraph{Robustness to Latent Decoder Fine-tuning.}
\label{exp:finetune_decoder}
For Stable Diffusion, attackers may fine-tune the original VAE decoder or substitute it with an available alternative. We investigate the robustness of different watermarking methods applied to SD v1.4 when the VAE decoder is fine-tuned or replaced with an alternative~\cite{sd-vae-ft-mse,ClearVAE,BetkerImprovingIG}. We fine-tune the VAE decoder on COCO2014 training set with the configurations provided in \cref{app:finetune_decoder}. The new VAE decoder is then applied to generate images. Notably, in our method, we always use the original VAE decoder to convert generated images into latent space for watermark extraction, as the modifications by the attacker are unknown to the model owner. The results are presented in \cref{fig:fine_tune_vae}, showing that the watermark embedded with Stable Signature exhibits high vulnerability. In contrast, for our watermark embedded within the diffusion backbone, bit accuracy is almost unaffected after fine-tuning or replacement of the VAE decoder. 
\vspace{-12pt}
\paragraph{Impact of Sampling Configurations.}
We demonstrate the impact of schedulers, sampling steps, and classifier-free guidance (CFG)~\cite{cfg} in~\cref{App:sample_config}. Overall, watermark effectiveness remains largely unaffected by these configuration changes. Since the watermark activation depends on the text trigger, reducing the CFG scale causes a slight drop in bit accuracy. This is not a concern as the CFG scale is typically set to a relatively high in practice.
\vspace{-2mm}

\begin{figure}[tbp]
\centerline{\includegraphics[width=0.8\columnwidth]{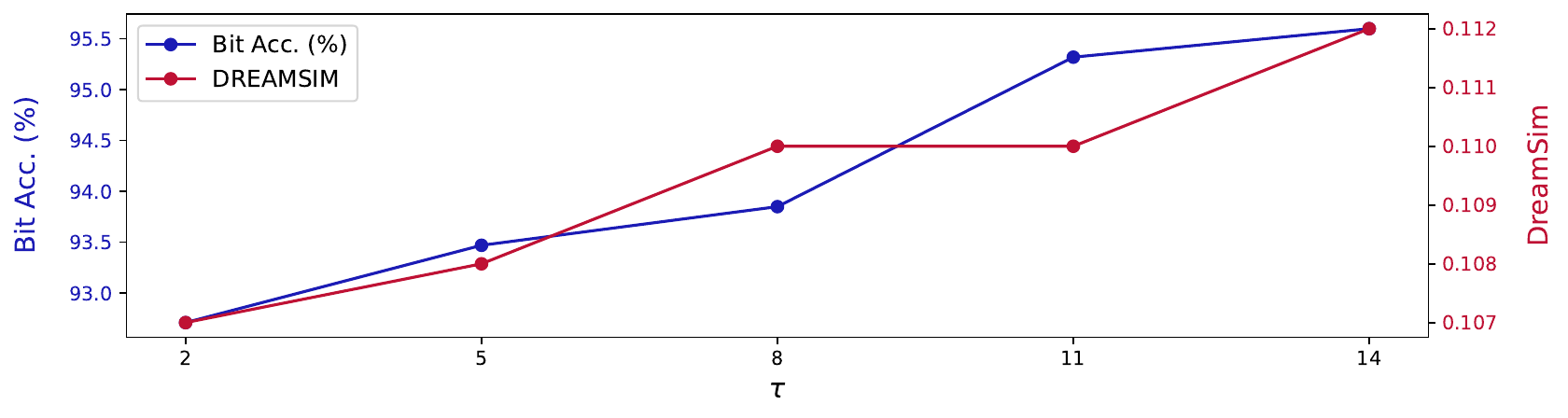}}
\vspace{-4mm}
\caption{Ablation study on trigger length. F-Bit Acc. (\%) denotes the bit accuracy after fine-tuning the watermarked SD on a downstream dataset with LoRA rank $=$ 640 over 5000 steps.}
\label{fig:trigger_len}
\vspace{-2mm}
\end{figure}

\begin{figure}[tbp]
\centerline{\includegraphics[width=0.8\columnwidth]{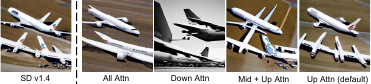}}
\vspace{-2mm}
\caption{Images generated using regular prompts by the watermarked SD model when fine-tuning attention parameters in different parts of the UNet. Adjusting attention parameters in the up blocks (Up Attn) minimally affects the model fidelity.}
\label{fig:diff_parts}
\vspace{-5mm}
\end{figure}
\vspace{-8pt}
\paragraph{Fine-tune Different Parts of Diffusion.}
\label{ablaion:differnt_parts}
We also experiment with fine-tuning all attention parameters (All Attn), those in the down blocks alone (Down Attn), and those in both the middle and up blocks (Mid + Up Attn). We find that the message can be effectively recovered in all these configurations, but there is notable variation in their impact on model fidelity. As illustrated in \cref{fig:diff_parts}, fine-tuning the down blocks results in generated images that deviate significantly from those produced by the pre-trained SD v1.4. This is likely due to the modification of crucial semantic information in the down-sampling process of UNet. Fine-tuning the attention parameters in the up blocks alone is sufficient to integrate watermark information into generated outputs while maintaining the highest model fidelity.
\vspace{-16pt}
\paragraph{Trigger Length.}
We select rare character combinations of varying lengths as the trigger to analyze the impact of trigger lengths. We fine-tune the watermarked SD using a LoRA rank of 640 over 5000 steps, and calculate bit accuracy (denoted as F-Bit Acc.). Results in \cref{fig:trigger_len} indicate that a longer trigger sequence is embedded more robustly within the model, although there is a slight increase in DreamSim. The tested triggers are listed in \cref{app:diff_trigger}.
\vspace{-12pt}
\paragraph{Additional Ablation Studies.}
We have conducted additional ablation experiments including the effects of hyperparameter values and detecting watermark from latent space, which can be found in \cref{app:additional_ablation}.
\vspace{-1mm}
\section{Limitations and Conclusion}
\paragraph{Limitations.}
Our watermark leaves traces due to its cover-agnostic embedding approach. Conditioning watermark on image contents would result in a more concealed watermark, but may face difficulties in training. Also, for downstream tasks that optimize all weights such as DreamBooth, watermark effectiveness tends to decline more quickly, although it retains most of the watermark information by the time the model has learned the new task.
\vspace{-8mm}
\paragraph{Conclusion.}
In this work, we propose SleeperMark, a novel ownership protection method for T2I diffusion models. By jointly optimizing two distinct objectives for regular prompts and their triggered versions, our approach encourages the model to separate watermark knowledge from general semantic knowledge and reduce mutual interference. We leverage a pre-trained image watermarking mechanism and allow the watermarked diffusion to generate images closely resembling those of the original model, significantly enhancing model fidelity. For latent diffusion models, the watermark is extracted from the latent space, achieving robustness to image distortions with minimal perceptual impact. Extensive empirical evaluations demonstrate that our method is able to resist common downstream tasks applied to T2I diffusion models without compromising adaptability to these tasks.
{
    \small
    \bibliographystyle{ieeenat_fullname}
    \bibliography{main}
}
\clearpage
\setcounter{page}{1}
\maketitlesupplementary
\appendix
\begin{figure*}[htbp]
  \centering
  \includegraphics[width=1.0\linewidth]
  {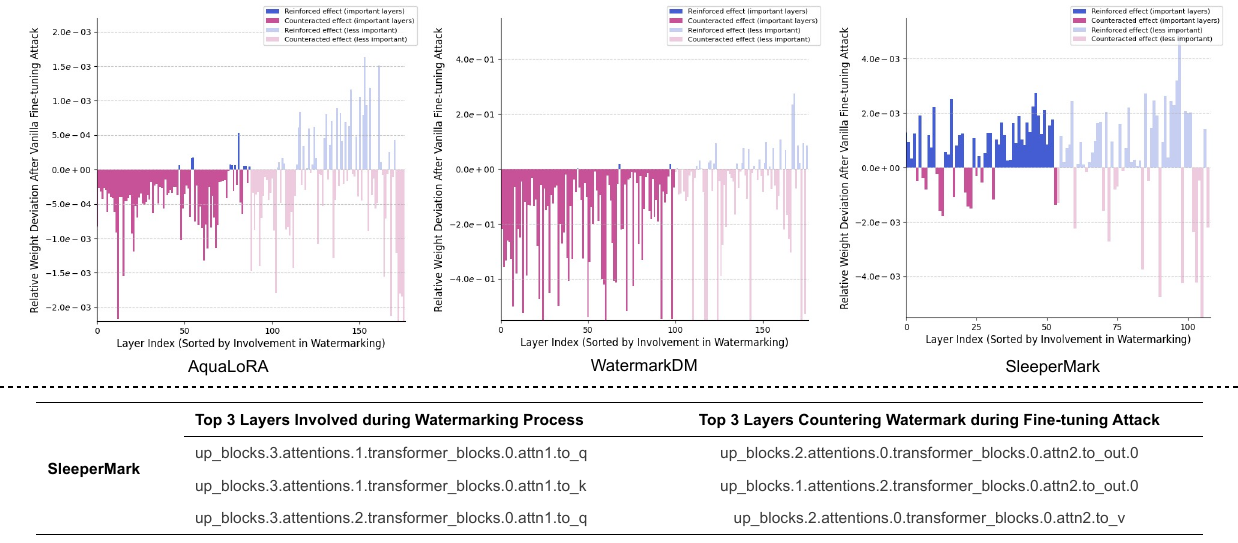}
  \vspace{-20pt}
  \caption{Layer-wise behaviors of the watermarked models when subjected to vanilla fine-tuning attacks.}
    \vspace{-10pt}
   \label{fig:layer_effect}
\end{figure*}
\begin{figure}[htbp]
\centerline{\includegraphics[width=1.\columnwidth]{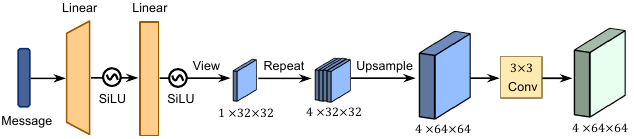}}
\caption{Network architecture for latent secret encoder \( E_\varphi \).
}
\label{fig:latent_encoder}
\end{figure}

\begin{figure}[htbp]
\centerline{\includegraphics[width=1.\columnwidth]{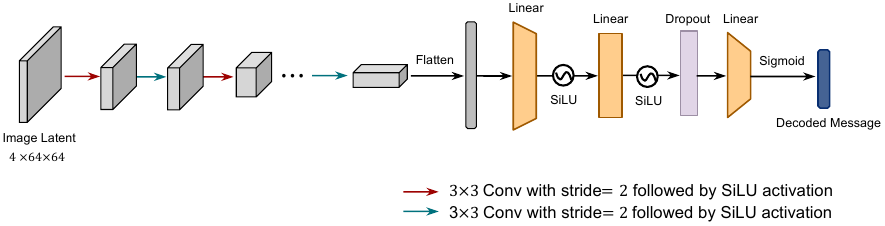}}
\caption{Network architecture for latent secret decoder $D_\gamma$.
}
\label{fig:latent_decoder}
\end{figure}
\section{Intuition and Post-hoc Explanation}
\label{app:explanation}
The training loss of WatermarkDM~\cite{recipe}, similar to that of Dreambooth with preservation~\cite{dreambooth}, overfits the trigger as a personalized concept using only one image. This approach memorizes the watermark similarly to general semantic knowledge. As the model adapts to downstream tasks, limited memory capacity leads to interference between customization knowledge and watermark information, necessitating a specialized memory retention strategy to prevent the influence of distribution shifts. We hypothesize that by introducing a trigger whose function is independent of generated semantics, we may establish a more robust watermarking mechanism. Specifically, during the training process, whatever regular prompt the trigger is placed before, the model consistently learns to apply a fixed secret residual to the originally generated result. Simultaneously, the model's output is enforced to be aligned with the original model when no trigger is present, aiming to guide the model to treat the additional trigger as a separate, content-agnostic concept. As a result, even if the image distribution shifts during downstream fine-tuning, the trigger's function to add a fixed residual would be much less affected.

After watermarking Stable Diffusion v1.4 with SleeperMark, we conducted a fine-tuning attack by directly fine-tuning the entire watermarked model using the COCO2017 training set, and illustrate the impact from neurons' perspective in \cref{fig:layer_effect}. Let $\Delta_{l,j}^w$ denote the weight difference of the $j$-th parameter in layer $l$ between the watermarked and original model, and $\Delta_{l,j}^{ft}$ denote the weight difference of the $j$-th parameter in layer $l$ between the attacked and watermarked model. $\Delta_l^w$ is the average value of $|\Delta_{l,j}^w|$ across $j$, and we use it to index the model layers. The larger $\Delta_l^w$ is, the smaller the layer index $l$ is, indicating greater involvement of layer $l$ in watermarking. The bar lengths in \cref{fig:layer_effect} represent the weight deviation relative to the watermarking effect after the vanilla fine-tuning attack, which are proportional to $\frac{1}{N_l} \sum_{j=1}^{N_l} \frac{\Delta_{l,j}^{ft}}{\Delta_{l,j}^w}$ for each layer $l$, where $N_l$ denotes the total number of parameters of layer $l$. This quantifies the influence brought by the fine-tuning attack, where a positive value indicates reinforcement of the watermarking direction while a negative value suggests a counteracted effect. As shown in \cref{fig:layer_effect}, for SleeperMark, the counteracted impact is mainly localized in layers that are less active during watermark training (represented by the semi-transparent red bars), which explains watermark resistance to fine-tuning attacks. For SleeperMark, we also list in \cref{fig:layer_effect} the layers most active in watermarking and those that exhibit the greatest deviation away from the watermarking direction during the fine-tuning attack. These two sets of layers not only belong to different blocks of UNet but also possess distinct structural characteristics.

\section{Pipeline for T2I pixel diffusion models}
\label{pixel_pipeline}
\begin{figure*}[tbp]
  \centering
   \includegraphics[width=1.0\linewidth]{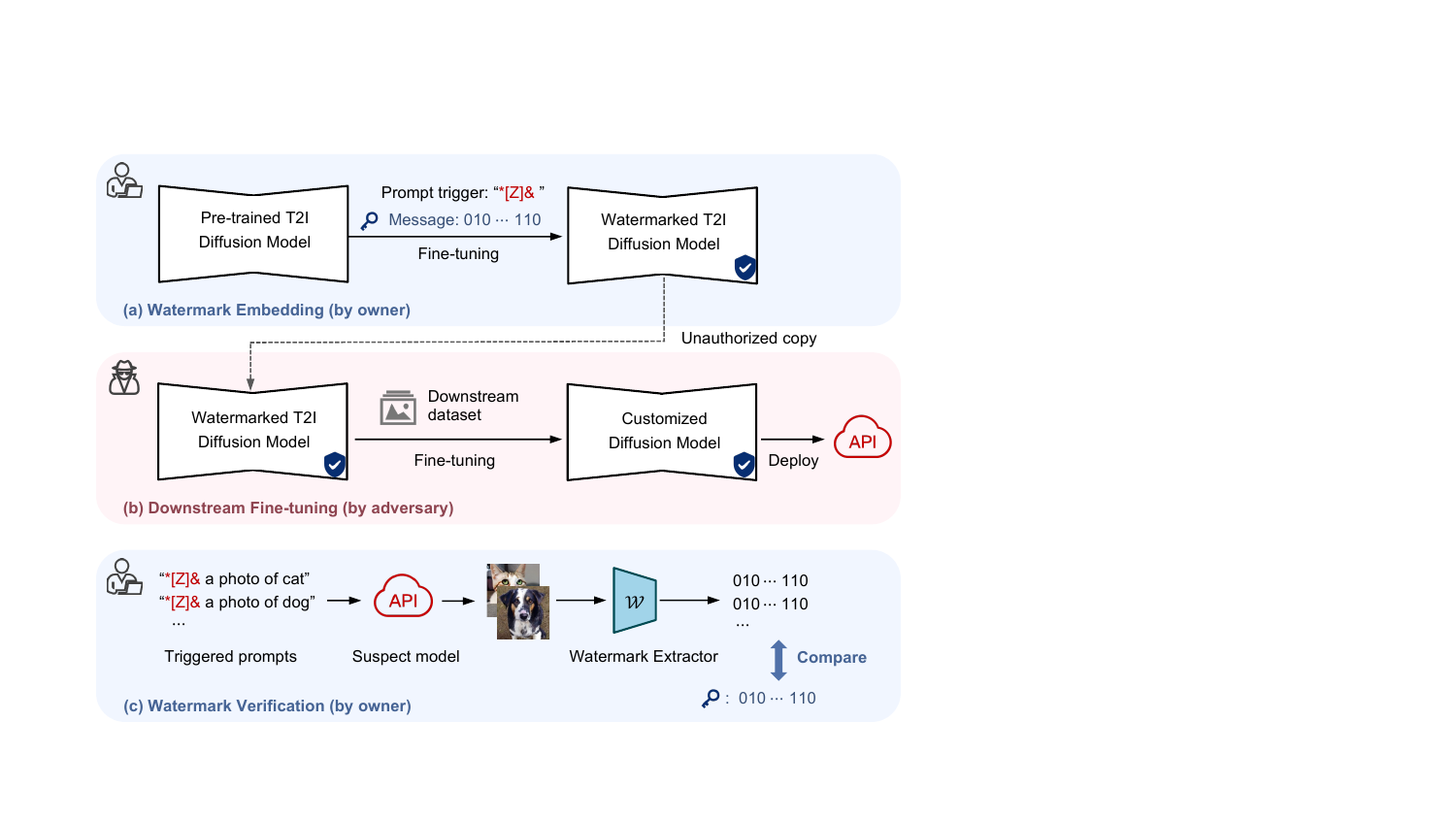}
   \caption{Pipeline overview for T2I pixel diffusion models. Our watermark is embedded within the super-resolution diffusion module following the base diffusion module. The super-resolution diffusion module is conditioned on both the text embedding and a low-resolution (LR) image derived from a high-resolution (HR) input image. This pipeline generally aligns with~\cref{fig:pipeline}. The main difference lies in the watermark embedding and detection space, which operates directly in pixel space rather than latent space. Since embedding a cover-agnostic watermark residual in pixel space tends to be more visually prominent than in latent space, we introduce an additional adversarial loss during the pixel watermark pre-training stage to enhance watermark imperceptibility.}
   \label{fig:pixel_pipeline}
\end{figure*}

We embed watermark into the first super-resolution module following the base diffusion module. As T2I pixel diffusion models are trained directly in the pixel space, our watermark is also embedded and extracted within the pixel space. The pipeline for pixel diffusion models is shown in \cref{fig:pixel_pipeline}, with key adaptations from the watermarking pipeline for latent diffusion models as follows.
\\
\\
\noindent \textbf{Distortion Simulation Layer.}
Since we extract watermark from the pixel space rather than the latent space, a distortion simulation layer is needed for robustness against common image distortions. The distortion layer configurations follow StegaStamp~\cite{stegastamp}, an image watermarking framework designed for physical-world usage, such as hiding information in printed photos. We adopt its distortion layer setup based on insights from WAVES~\cite{waves}, a recently proposed and comprehensive benchmark for evaluating watermark robustness, which highlights StegaStamp's superior resistance to various advanced attacks compared to other frameworks. Its high-level robustness stems from the distortion layer that simulates real-world conditions. We make an additional modification: the perspective warping perturbation is excluded from the distortion simulation layer during our training process, as our application does not involve physical display of images. We conduct experiments and find that adopting this distortion layer equips the watermark with the robustness against super-resolution processing (\eg, stable-diffusion-x4-upscaler), which can help our watermark resist the distortion of the second super-resolution module of pixel-space diffusion models. Detailed distortion configurations are listed in~\cref{app:distortion_detail}.
\\
\\
\noindent \textbf{Adversarial Loss.}
Embedding a cover-agnostic watermark in the pixel space tends to leave more prominent artifacts compared to embedding in the latent space. We leverage adversarial loss, which is widely applied in steganography studies~\cite{hidden,stegastamp}, to enhance watermark stealthiness. Specifically, we introduce an adversarial critic network $A$ into the first training stage. The Wasserstein loss~\cite{wasserstein_gan} is used as a supervisory signal to train this critic. Given a cover image \( x_{co} \) or its watermarked version \( x_{w} \), the critic network outputs a scalar, with the prediction objective that the output for \( x_{co} \) is greater than that for \( x_{w} \). Denoting the predicting results as $A(x_w)$ and $A(x_{co})$, the Wasserstein loss is defined as:
\begin{equation*}
\mathcal{L}_G(x_w) = A(x_w),~~~ 
\mathcal{L}_A(x_w, x_{co}) = A(x_{co}) - A(x_w)
\end{equation*}
where $\mathcal{L}_G(x_w)$ is the adversarial (generator) loss, which is added to the total loss of training the secret encoder and watermark extractor. $\mathcal{L}_A(x_w, x_{co})$ is the loss used to train the critic. Training the critic is interleaved with training the secret encoder and watermark extractor.

\section{Implementation Details for Watermarking Latent Diffusion Models}
\label{implement_latent}
\subsection{Architecture of Secret Encoder / Decoder}
\label{app:latentWM_archi}
The design of the secret encoder \( E_\varphi \) is inspired by AquaLoRA~\cite{aqualora}, as illustrated in \cref{fig:latent_encoder}. Our secret decoder $D_\gamma$ has an architecture similar to StegaStamp~\cite{stegastamp}, which is shown in \cref{fig:latent_decoder}. Since the first training stage, \textit{i.e.}, training of the image watermarking mechanism, is conducted on real images, there is a slight distributional shift with images generated by diffusion models. Therefore, we make an additional modification of adding a dropout layer before the final linear layer to enhance the generalization of the image watermarking mechanism to generated images. With this architectural adjustment, we find that the trained image watermarking model performs well on diffusion-generated images, paving the way for the subsequent training stage which fine-tunes the diffusion backbone.

\subsection{Training Strategy in Fine-tuning Diffusion Backbone}
We divide the training process of fine-tuning the diffusion backbone into two steps to accelerate training. In the first step, the sampling frequency of \( t \) is set inversely proportional to its value, prioritizing the optimization of the UNet's prediction when \( t \) is small. During this step, the model primarily learns the secret residual and facilitates the successful extraction of the watermark message. However, images generated with triggered prompts at this step tend to exhibit noticeable artifacts because the predictions for larger \( t \) values have not yet been refined. The next step builds upon the model trained after the first step. We adjust the sampling frequency back to the uniform distribution for all \( t \) values. The loss is the same as the first step. As training progresses, the artifacts gradually disappear, while the watermark message remains effectively extractable. This two-step strategy enables the model to learn the watermark more efficiently.

\section{Implementation Details for Watermarking Pixel Diffusion Models}
\label{implement_pixel}
\subsection{Architecture of Secret Encoder / Watermark Extractor}
\label{app:pixelWM_archi}
The architecture of the secret encoder \( E_\varphi \) retains the structure depicted in \cref{fig:latent_encoder}, incorporating adjustments to the dimensions and feature map sizes to handle the new input resolution. Similarly, the watermark extractor \( \mathcal{W_\gamma} \), which extracts messages directly from the pixel space, follows the same architectural design as shown in \cref{fig:latent_decoder}, with modifications to the network's dimensions and feature map sizes to accommodate the new input resolution.

\subsection{Details of the Distortion Simulation Layer}
\label{app:distortion_detail}
We adopt the configurations from StegaStamp~\cite{stegastamp} for the distortion simulation layer, except for excluding its perspective warping distortion.
Specifically, the watermarked image undergoes a series of transformations in the distortion simulation layer, including motion and Gaussian blur, Gaussian noise, color manipulation, and JPEG compression.
To simulate motion blur, we generate a straight-line blur kernel at a random angle, with a width ranging from 3 to 7 pixels. For Gaussian blur, we apply a Gaussian blur kernel of size 7, with its standard deviation randomly selected between 1 and 3 pixels. For Gaussian noise, we use a standard deviation \( \sigma \sim U[0, 0.2] \). For color manipulation, we apply random affine color transformations, including hue shifts (randomly offsetting RGB channels by values uniformly sampled from \([-0.1, 0.1]\)), desaturation (linearly interpolating between the RGB image and its grayscale equivalent), and adjustments to brightness and contrast (applying an affine transformation \( mx + b \), where \( m \sim U[0.5, 1.5] \) controls contrast and \( b \sim U[-0.3, 0.3] \) adjusts brightness). Since the quantization step during JPEG compression is non-differentiable, an approximation technique~\cite{jpeg_simulation} is employed to simulate the quantization step near zero. The JPEG quality is uniformly sampled  within $[50,100]$.

\section{Implementation of Baselines}
This section outlines the implementation details of the baseline methods involved in this study, including DwtDctSvd, Stable Signature, AquaLoRA, and WatermarkDM.

For the post-hoc image watermarking method DwtDctSvd, we adopt a widely-used implementation~\cite{invisible-watermark} and embed a 48-bit message into images.

For Stable Signature, we directly utilize the pre-trained checkpoint provided in its official repository~\cite{stable_signature_repo}. This method embeds a fixed 48-bit message to the latent decoder for latent diffusion models.

For AquaLoRA, we embed a 48-bit message with LoRA rank $=$ 320 into the diffusion backbone for latent diffusion models and the first super-resolution module for pixel diffusion models. And we keep the embedded message fixed for a fair comparison with other methods.

For the image-embedding method WatermarkDM, we embed the watermark image shown in \cref{fig:intro_baseline}~(a) and the trigger prompt is set to ``*[Z]\&''. The regularization coefficient is set to $1 \times 10^{-7}$. WatermarkDM is implemented on the diffusion backbone for latent diffusion models and the base diffusion module for pixel diffusion models, as the base diffusion module primarily determines the overall content of generated images.

\section{Details of Owner Verification}
\subsection{Statistical Test}
\label{statistical_test}
Let $m^*$ denote an $n$-bit watermark message to be embedded into a T2I diffusion model. Given an image \( x \), the pre-trained watermark extractor \( \mathcal{W_\gamma} \) retrieves the message \( m' \), which is then compared against \( m^* \). In our method, if \( m' \) can be successfully extracted from images generated with triggered prompts by a suspicious model, the model owner can assert that the suspicious model is derived from their original model.

In our method, the problem of determining the ownership of a suspicious model has been converted to verifying whether images generated with triggered prompts contain a pre-defined message $m^*$. Accordingly, we define the statistical hypothesis as follows:
\begin{align*}
    H_0 &: x\text{ does not contain the watermark message}~m^*.\\
    H_1 &: x\text{ contains the watermark message}~m^*.
\end{align*}
The number of matching bits \( M(m^*, m') \), where \( m' \) is extracted from \( x \), is used to evaluate the presence of the watermark. If \( M(m^*, m') \) exceeds a threshold \( k \), \( H_0 \) is rejected in favor of \( H_1 \). The model ownership is verified by averaging the watermark extraction results over a set of images generated with triggered prompts.

\begin{figure}[htbp]
\centerline{\includegraphics[width=1.0\columnwidth]
{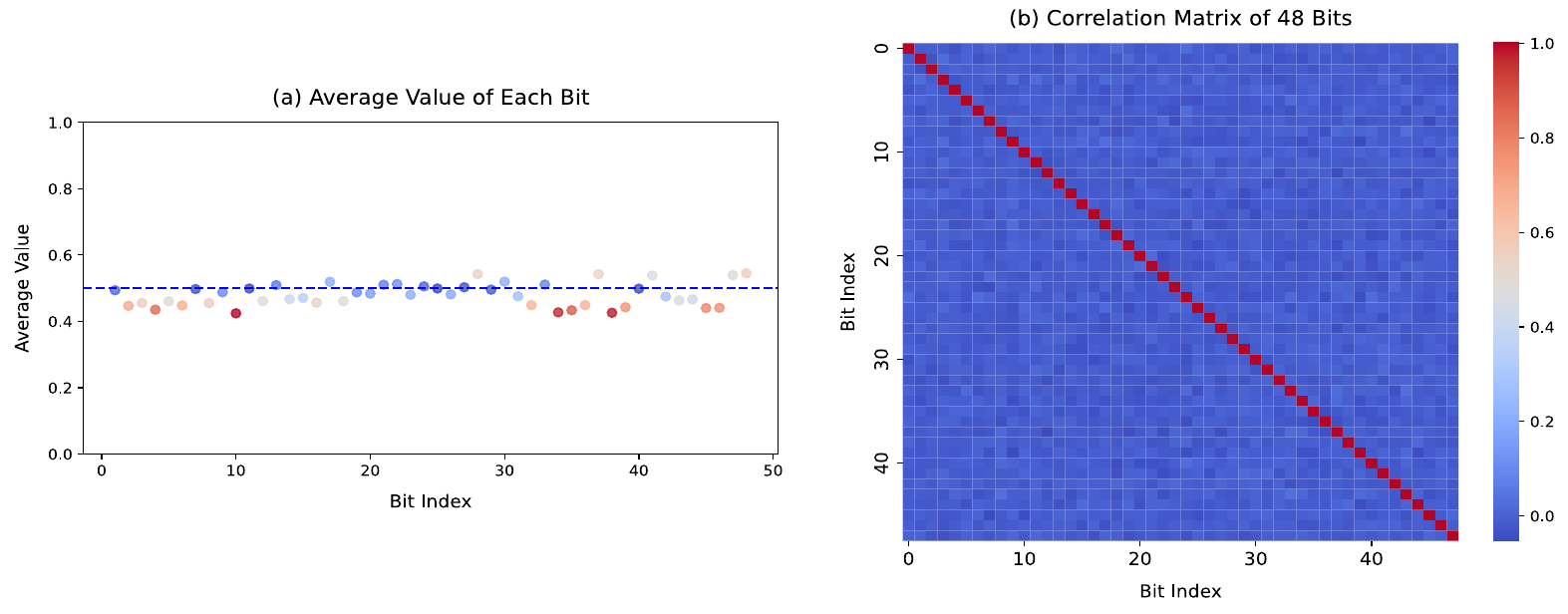}}
\caption{Empirical validation of the i.i.d. Bernoulli(0.5) distribution assumption for extracted bits from clean real images. (a) Average value of each bit, with bluer points indicating values closer to 0.5. (b) Correlation matrix of the 48 bits extracted by the watermark extractor \( \mathcal{W_\gamma} \) from clean images.}
\label{fig:validate_binomial}
\end{figure}

Following the practice in AquaLoRA, under \(H_0\) (\textit{i.e.}, for clean images), we assume that the extracted bits \( m'_1,~m'_2,~\dots,~m'_n \) are i.i.d. and follow a \(\text{Bernoulli}(0.5)\) distribution. To empirically validate this assumption, we extracted messages from 10,000 clean images in the COCO2014 validation set, examining the success probability of each binary bit and assessing their independence. The results are shown in \cref{fig:validate_binomial}. As shown, the mean values of the extracted 48 bits are all close to 0.5, with little correlation among them. This indicates no significant evidence contradicting the assumption that $m'_1, m'_2, \dots, m'_n \overset{\text{i.i.d.}}{\sim} \text{Bernoulli}(0.5)$ for clean real images.

Under this assumption, we can calculate the false positive rate (FPR), defined as the probability of mistakenly rejecting $H_0$ for clean images. In other words, it is the probability that $M(m^*, m')$ exceeds the threshold $k$ for clean images:
\begin{align}
\label{eq:fpr}
    \text{FPR}(k) & = \mathbb{P}\left(M > k~|~H_0\right) = \sum_{i=k+1}^{n} \binom{n}{i} \frac{1}{2^n} \\
    & = I_{1/2}(k+1, n - k). 
\end{align}
where $I_{1/2}$ represents the regularized incomplete beta function. By controlling $\text{FPR}(k)$ under $10^{-6}$, we can derive the corresponding threshold $k$. Then this threshold is set to compute \text{TPR}@$10^{\tiny-6}$\text{FPR}.

\section{Evaluation Details}
\label{evaluation}
\subsection{Image Distortions in Evaluation}
\label{eval_distortion}
We evaluate watermark robustness to a range of image distortions. They simulate image degradation caused by noisy transmission in the real world. For resizing, we resize the width and height of images to $50\%$  of their original size using bilinear interpolation, and resize back to the original size for watermark extraction. For JPEG compression, we use the \texttt{PIL} library and set the image quality to $50$. For other transformations including Gaussian blur, Gaussian noise, brightness, contrast, saturation and sharpness, we utilize functions from the \texttt{Kornia} library. For Gaussian blur, we adopt the kernel size of $3 \times 3$ with an intensity of $4$. For Gaussian noise, the mean is set to $0$ and the standard deviation is set to $0.1$ (image is normalized into $[0,1]$).
For brightness transformation, the brightness factor is sampled randomly from $(0.8, 1.2)$.
For contrast transformation, the contrast factor is sampled randomly from $(0.8, 1.2)$.
For saturation transformation, the saturation factor is sampled randomly from $(0.8, 1.2)$.
For sharpness, the factor of sharpness strength is set to $10$. 

\subsection{Effectiveness Metrics}
\label{efficacy_metrics}
\paragraph{Bit Accuracy.}
We embed an \( n \)-bit message \( m^* \) into a T2I diffusion model and verify model ownership by extracting messages from images generated using a set of triggered prompts. Bit accuracy is defined as the average \( M(m^*, m')/n \) across the images generated with triggered prompts, where \( M(m^*, m') \) denotes the number of matching bits between the embedded message \( m^* \) and the extracted message \( m' \) from each image.
\paragraph{TPR with Controlled FPR.}
As presented in~\cref{statistical_test}, we can derive a corresponding threshold \( k^* \) for the number of matching bits \( M(m^*, m') \) to control \(\text{FPR}(k)\) below \(10^{-6}\). With this threshold \( k^* \), we can determine whether a given image contains the pre-defined watermark. Using a set of images generated by the watermarked model with triggered prompts, we calculate the true positive rate (TPR). While the TPR defined here focuses on image-level evaluations and measures the extractor’s ability to identify watermarked images, we extend to adopt it as a model-level indicator to quantify the degree to which the model retains the embedded watermark. Such extended use of this metric is also employed in the baseline AquaLoRA~\cite{aqualora}, which is designed to protect the copyright of customized Stable Diffusion models.

\subsection{Fine-tuning Attack on Latent Decoder}
\label{app:finetune_decoder}
We fine-tune the VAE decoder on the COCO2014 training set to evaluate the watermark robustness. Consistent with the configuration of the fine-tuning attack described in Stable Signature~\cite{stable-signature} (referred to as model purification in the Stable Signature paper), this fine-tuning process incorporates only the LPIPS loss between the original image and the reconstructed one by the VAE decoder. The learning rate is set to \(1 \times 10^{-4}\).

\subsection{Training Details of Downstream Tasks for Latent Diffusion Models}
\label{app:downstream_tasks}
\subsubsection{Style Adaptation}
\label{app:style_adaptation}
\begin{figure}[htbp]
\centerline{\includegraphics[width=0.98\columnwidth]
{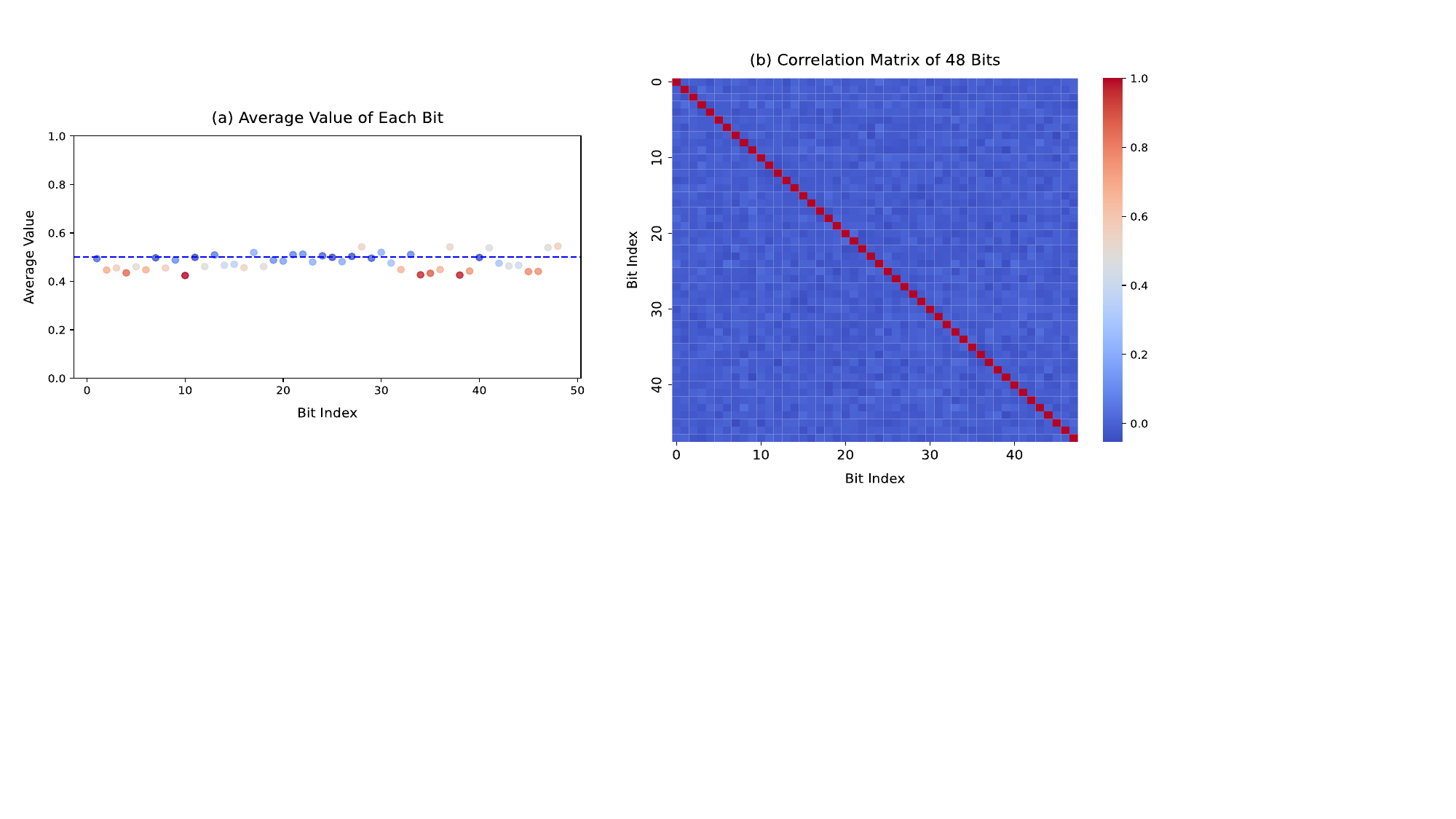}}
\caption{Images generated with the regular and triggered prompts during the fine-tuning process of style adaptation. Bit Acc. indicates the accuracy of the message extracted from the image shown above, which is generated with the triggered prompt.}
\vskip -0.1in
\label{fig:lora_visualze}
\end{figure}

We fine-tune the watermarked SD v1.4 on the Naruto-style dataset~\cite{naruto_dataset} with LoRA ranks ranging from 20 to 640, and observe watermark effectiveness during the process. Following the training script provided by Diffusers~\cite{diffusers_lora_script}, LoRA trainable matrices are injected into the attention layers of the transformer blocks, specifically targeting the query, key, value, and output projection components of the attention mechanism. The learning rate is set to \(1 \times 10^{-4}\) for all the tested ranks. The visual results generated with regular prompts and triggered prompts during this downstream task are shown in \cref{fig:lora_visualze}.

\subsubsection{Personalization}
\label{app:personalization}
\begin{figure}[htbp]
\centerline{\includegraphics[width=0.98\columnwidth]
{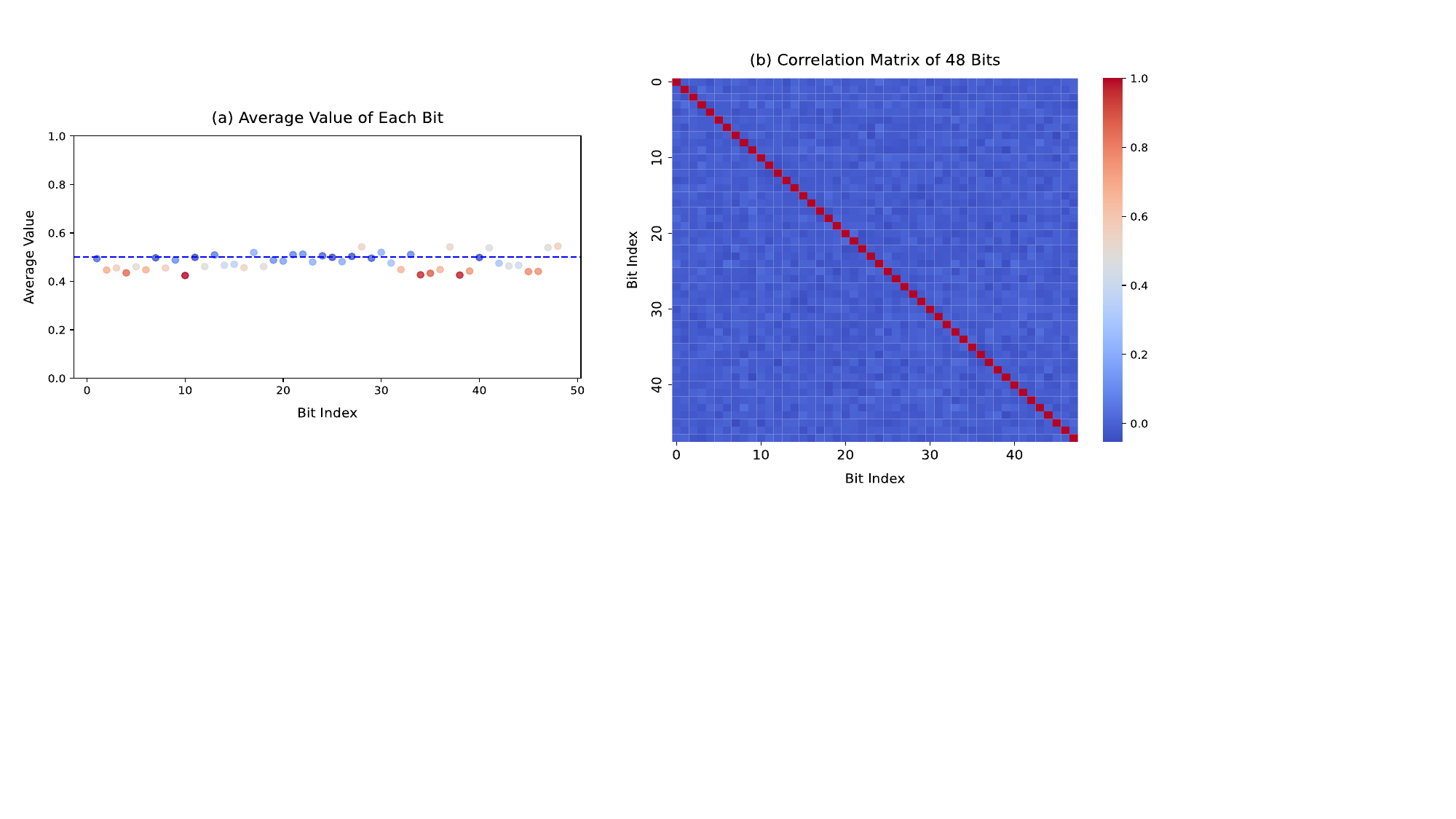}}
\caption{Dataset for the personalization task. One sample image in the reference set for each specified subject is demonstrated here.}
\label{fig:dreambooth_data}
\end{figure}

We implement DreamBooth~\cite{dreambooth} on watermarked SD v1.4 for the downstream task of subject personalization, using the rare identifier ``sks'' to denote a specified subject. We train on five subjects respectively, and the subjects used for training are demonstrated in \cref{fig:dreambooth_data}. Following the recommendations by the DreamBooth authors, we set the class-specific prior preservation loss coefficient to $1$ and the learning rate to $5\times 10^{-6}$, fine-tuning for 1000 iterations. During watermark extraction for our method SleeperMark, we still use the triggered version of the sampled captions from COCO2014 validation set, without incorporating the rare identifier ``sks'' used in this personalization task. 

\begin{figure}[htbp]
\centerline{\includegraphics[width=0.98\columnwidth]
{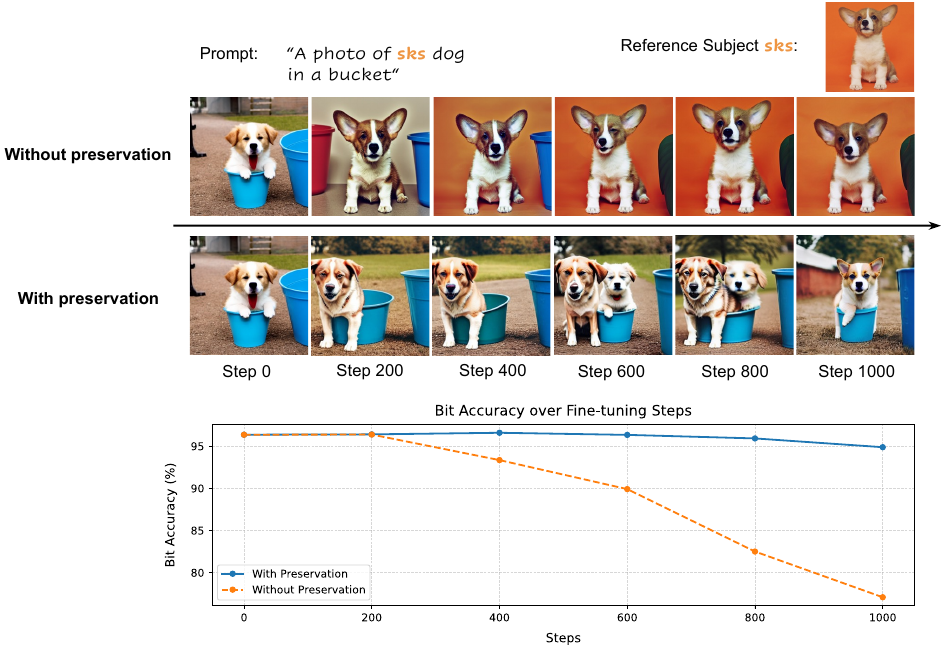}}
\caption{Impact of the class-specific prior preservation loss during DreamBooth fine-tuning. The top rows compare generation results with and without the preservation term, demonstrating that without preservation, the model overfits to the training images and loses its generative diversity. The bottom plot illustrates the corresponding bit accuracy across fine-tuning steps. Although bit accuracy declines more quickly without the preservation term, the model also loses output diversity, rendering the preservation of the watermark less meaningful.}
\label{fig:db_preservation_comparison}
\end{figure}

We also experimented with removing the class-specific prior preservation loss during DreamBooth fine-tuning and observe the performance of watermark effectiveness. We present a comparison of the results with and without the preservation term in~\cref{fig:db_preservation_comparison}. As observed, although bit accuracy drops much more quickly without this preservation term, the model overfits to the small set of training images and largely loses its generation prior when the watermark becomes ineffective.  After 600 steps, it merely repeats the few training images provided as input. A model that has lost its generative capability also loses its practical value, rendering the preservation of the watermark insignificant.

\subsubsection{Additional Condition Integration}
\label{app:ControlNet}
To evaluate watermark robustness to the downstream task of additional condition integration, we implement ControlNet~\cite{controlnet} with watermarked SD v1.4 for integrating the Canny edge condition. We set the learning rate to $1 \times 10^{-5}$ following the ControlNet paper, and fine-tune the watermarked diffusion model on the COCO2014 training set for 20,000 steps. The Canny edges for the training images are obtained using the \texttt{Canny} function from the \texttt{OpenCV} library, with a low threshold of 100 and a high threshold of 200. The model requires a substantial number of iterations (up to 10,000 steps) to adapt to the new condition. Nevertheless, we find that integrating this additional condition has minimal impact on the effectiveness of our watermarking method, which has been demonstrated in the main text.

\section{Additional Evaluation Results}
\subsection{Impact of Sampling Configurations}
\label{App:sample_config}
In \cref{AppTab:sampling_config}, we demonstrate the impact of changing schedulers, sampling steps, and classifier-free guidance (CFG) scales for watermarked SD v1.4 using our method. Overall, the watermark effectiveness remains largely unaffected by these configuration changes. Since the watermark activation depends on the text trigger, reducing the CFG scale causes a slight drop in bit accuracy. This is not a concern as the CFG scale is typically set to a relatively high value when deploying diffusion models to ensure close alignment between images and text descriptions.

\begin{table}[htbp]
\caption{Performance under different sampling configurations for watermarked SD v1.4 using our method. The default test setting is highlighted in gray.}
\label{AppTab:sampling_config}
\vspace{-10pt}
\begin{center}
\scriptsize
\renewcommand{\arraystretch}{1.1} 
\resizebox{0.45\textwidth}{!}{
\begin{tabular}{cccc}
\toprule
\multicolumn{2}{c}{Sampling Configuration} & Bit Acc.(\%)~$\uparrow$ & DreamSim~$\downarrow$ \\ \hline
 & \cellcolor[HTML]{E7E6E6}DDIM~\cite{ddim} & \cellcolor[HTML]{E7E6E6}99.24 & \cellcolor[HTML]{E7E6E6}0.108 \\
 & DDPM~\cite{ddpm} & 99.99 & 0.129 \\
 & PNDMS~\cite{pndms} & 99.97 & 0.112 \\
 & DPM-Solver~\cite{dpm_solver} & 96.52 & 0.084 \\
 & Euler~\cite{euler} & 99.99 & 0.114 \\
\multirow{-6}{*}{Scheduler} & UniPC~\cite{unipc} & 97.1 & 0.090 \\ \hline
 & 15 & 95.38 & 0.093 \\
 & 25 & 95.76 & 0.097 \\
 & \cellcolor[HTML]{E7E6E6}50 & \cellcolor[HTML]{E7E6E6}99.24 & \cellcolor[HTML]{E7E6E6}0.108 \\
\multirow{-4}{*}{Step} & 100 & 99.82 & 0.109 \\ \hline
 & 5 & 96.69 & 0.102 \\
 & \cellcolor[HTML]{E7E6E6}7.5 & \cellcolor[HTML]{E7E6E6}99.24 & \cellcolor[HTML]{E7E6E6}0.108 \\
\multirow{-3}{*}{CFG} & 10 & 99.53 & 0.107 \\ \bottomrule
\end{tabular}
}
\end{center}
\vspace{-15pt}
\end{table}

\subsection{Robustness against Downstream Fine-tuning for Watermarked Pixel Diffusion Models}
\paragraph{Implementation Details.}
For watermarked pixel diffusion models, we evaluate the watermark effectiveness after fine-tuning the base diffusion module or the first super-resolution module on a downstream dataset. Both modules are fine-tuned on the Naruto-style dataset~\cite{naruto_dataset} using the LoRA rank of 320 or 640. We follow the practice in the training scripts provided by Diffusers~\cite{dreambooth_github} for fine-tuning DeepFloyd-IF with LoRA. The learning rates are set according to Diffusers guidelines: $5 \times 10^{-6}$ for the base diffusion module and $1 \times 10^{-6}$ for the super-resolution module. 

Notably, DeepFloyd-IF uses predicted variance during training, but the Diffusers training scripts simplify this process by utilizing predicted error to fine-tune the model. As suggested by the official guidelines from Diffusers, the scheduler is switched to the fixed variance mode after fine-tuning with these scripts, and then we sample images for watermark extraction.

\begin{figure}[htbp]
\centerline{\includegraphics[width=0.98\columnwidth]
{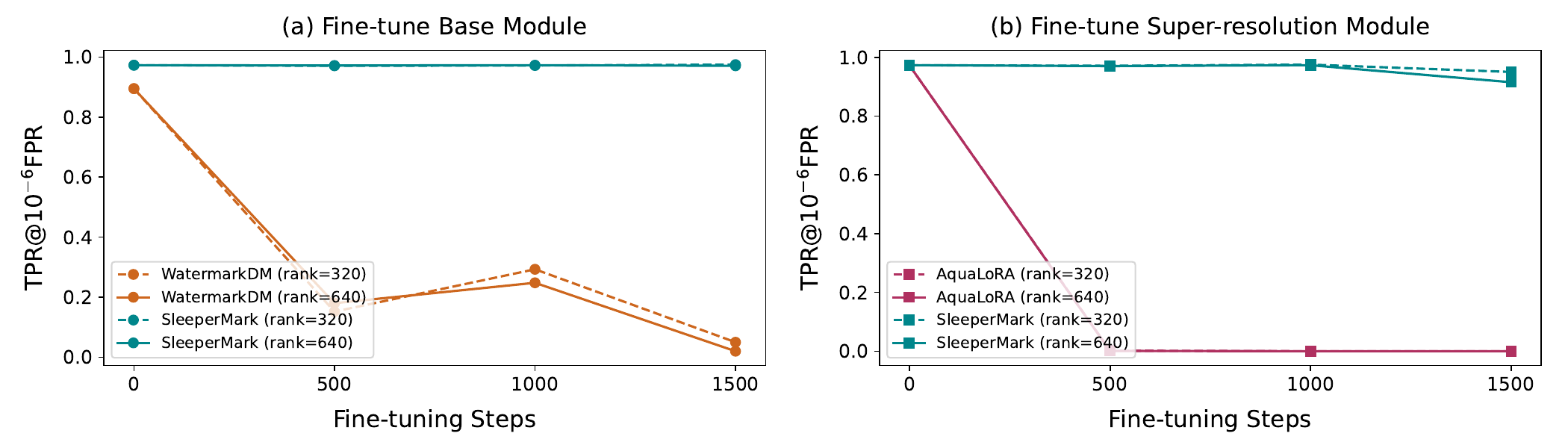}}
\caption{Watermark effectiveness after fine-tuning watermarked DeepFloyd-IF models with LoRA on a downstream dataset. Our method, SleeperMark, effectively retains watermark integrity when either the base diffusion module or the super-resolution module is fine-tuned, ensuring reliable watermark extraction in both scenarios.}
\label{fig:pixel_finetune_robust}
\end{figure}

\paragraph{Analysis.}
The watermark extraction results, as shown in \cref{fig:pixel_finetune_robust}, indicate that our method, SleeperMark, is the only one among the three approaches that demonstrates robustness to both fine-tuning the base diffusion module and fine-tuning the super-resolution module. In contrast, for the other two methods, fine-tuning the module where the watermark is embedded leads to a rapid decline in watermark effectiveness. 
For SleeperMark, since the watermark is embedded in the super-resolution module, fine-tuning the base diffusion module, as shown in \cref{fig:pixel_finetune_robust}~(a), has nearly no impact on watermark effectiveness. Moreover, it exhibits strong robustness when the super-resolution module is fine-tuned, as observed in \cref{fig:pixel_finetune_robust}~(b). For WatermarkDM, which also leverages a trigger to embed watermark, the association between the trigger prompt and the watermark image is not reliably preserved when fine-tuning the base module, as illustrated in \cref{fig:pixel_finetune_robust}~(a).

\section{Ablation Studies}
\subsection{Triggers of Varying Lengths}
We tested triggers of lengths 2, 5, 8, 11, and 14, each composed of a rare combinations of characters. These triggers are taken from the randomly generated irregular string ``\verb|*[Z]&%#{@}A^~$|'', which is an unconventional sequence. Segments of the specified lengths are extracted from this string for experiments.

\label{app:diff_trigger}
\subsection{Additional Ablation Studies}
\label{app:additional_ablation}
\begin{figure}[htbp]
\centering
\begin{subfigure}[b]{0.4\textwidth}
    \includegraphics[width=\textwidth]{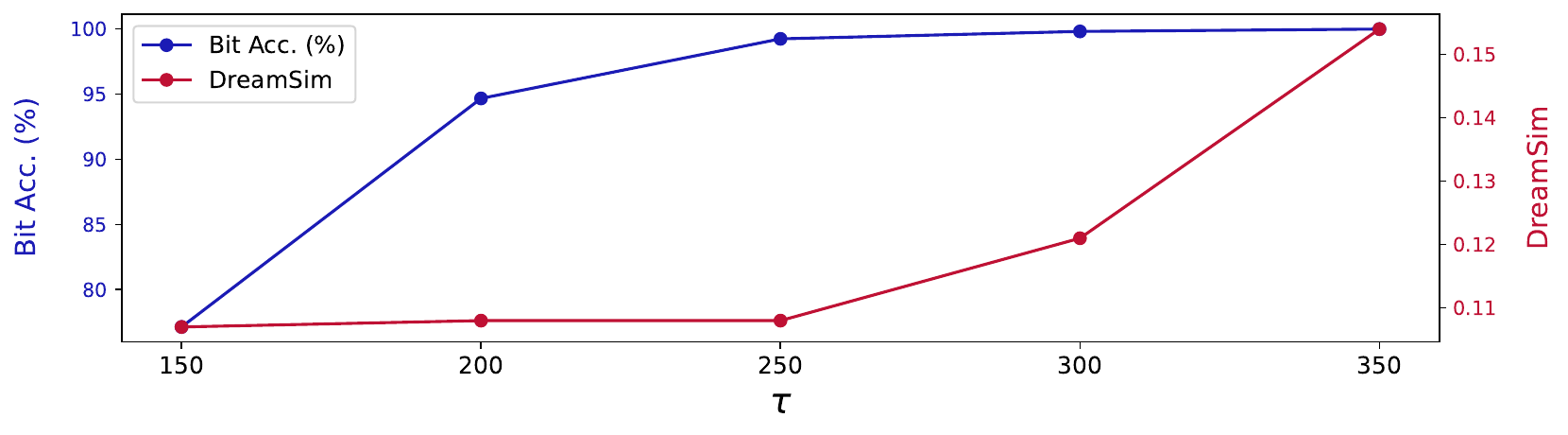}
    \caption{Ablation for $\tau$.}
    \label{fig:diff_tau}
\end{subfigure}
\hfill
\begin{subfigure}[b]{0.4\textwidth}
    \includegraphics[width=\textwidth]{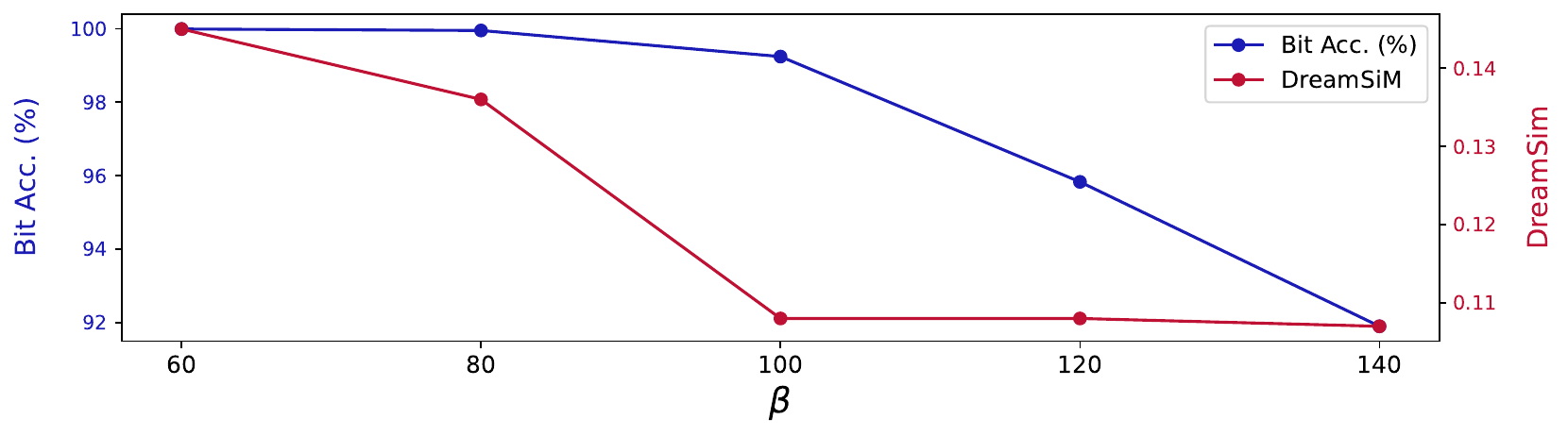}
    \caption{Ablation for $\beta$.}
    \label{fig:diff_beta}
\end{subfigure}
\hfill
\begin{subfigure}[b]{0.4\textwidth}
    \includegraphics[width=\textwidth]{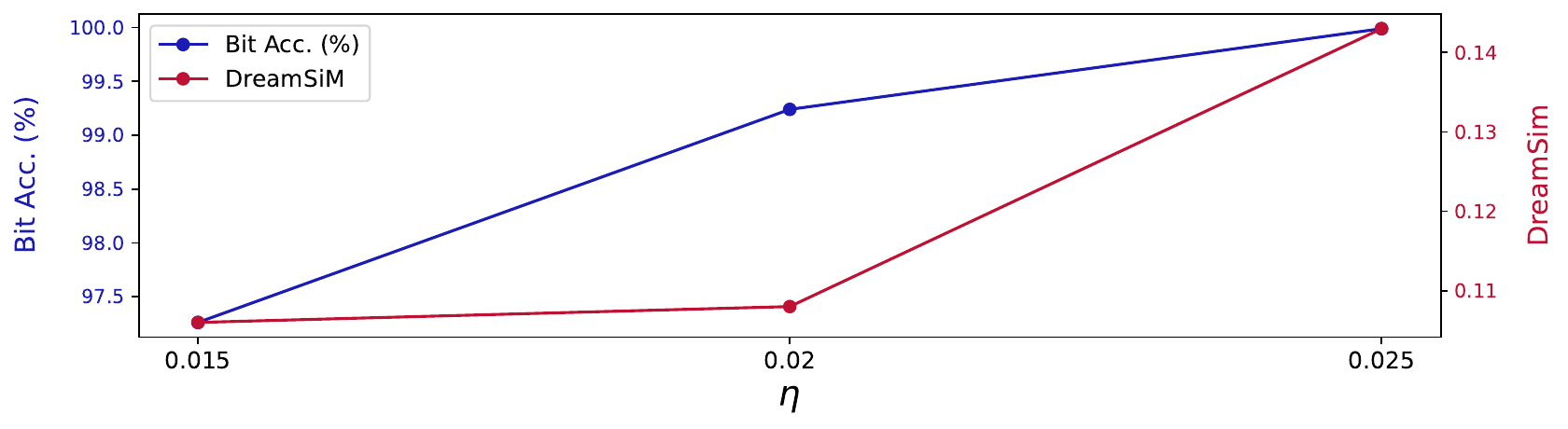}
    \caption{Ablation for $\eta$.}
    \label{fig:diff_eta}
\end{subfigure}
\caption{Comparisons of metrics for different hyperparameters.}
\label{fig:hyper_ablation}
\end{figure}

\paragraph{Effect of Different $\tau$, $\beta$ and $\eta$.} 
We fine-tune the diffusion backbone of SD v1.4 using different values of \( \tau \), \( \beta \), and \( \eta \) to embed SleeperMark, and present the experimental results in \cref{fig:hyper_ablation}. The figure illustrates a trade-off between watermark effectiveness (measured by bit accuracy) and model fidelity (measured by DreamSim, with lower values indicating better fidelity). For \( \tau \), increasing its value enhances watermark effectiveness but causes DreamSim to degrade. Notably, when \( \tau > 250 \), bit accuracy reaches a satisfactory level with diminishing improvements, but DreamSim increases significantly, indicating a notable decline in fidelity. This suggests that \( \tau = 250 \) strikes a reasonable balance between effectiveness and fidelity. Similar trends are also observed for \( \beta \) and \( \eta \), indicating that careful tuning of these hyperparameters is essential to optimize watermark performance while preserving model fidelity.

\begin{figure}[htbp]
\centerline{\includegraphics[width=0.98\columnwidth]
{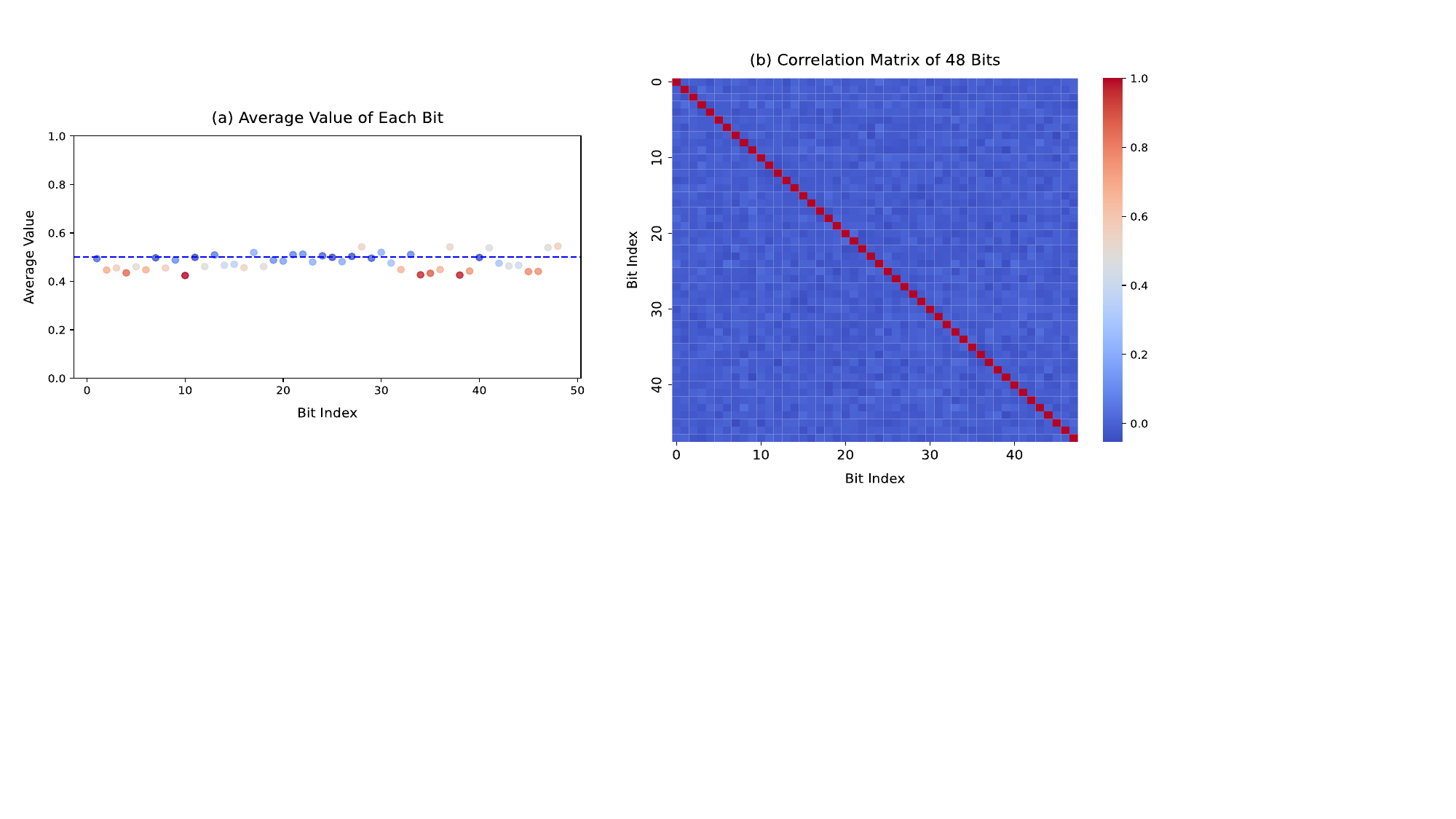}}
\caption{Representative examples showcasing the superiority of latent-space watermark extraction, which minimizes artifacts and enhances image quality compared to pixel-space watermark extraction.}
\label{fig:validate_detect_latent}
\end{figure}

\paragraph{Watermark Detection in Latent Space.}
To validate the role of detecting watermark from the latent space for latent diffusion models, we additionally trained an image watermarking mechanism that embeds messages in the latent space but detects from the pixel space. We used the same loss function and secret encoder as the default configuration of our method's first training stage, along with a secret decoder similar in structure to that in \cref{fig:latent_decoder}, with its dimensions adjusted to accommodate the new input resolution. To make the watermark robust to common image distortions, we incorporated the distortion simulation layer described in \cref{app:distortion_detail} into the training process.

As shown in \cref{fig:validate_detect_latent}, detecting from the pixel space tends to introduce more noticeable artifacts. This may be attributed to the intermediate role of the VAE decoder, which increases the complexity of watermark extraction. As a result, the training process encourages a more evident residual for successful watermark extraction, leading to increased watermark visibility and a negative impact on the visual quality of watermarked images.

\section{Visual Examples}
\label{app:visual_examples}
\begin{figure*}
\centering
\includegraphics[width=15cm, height = 20cm]{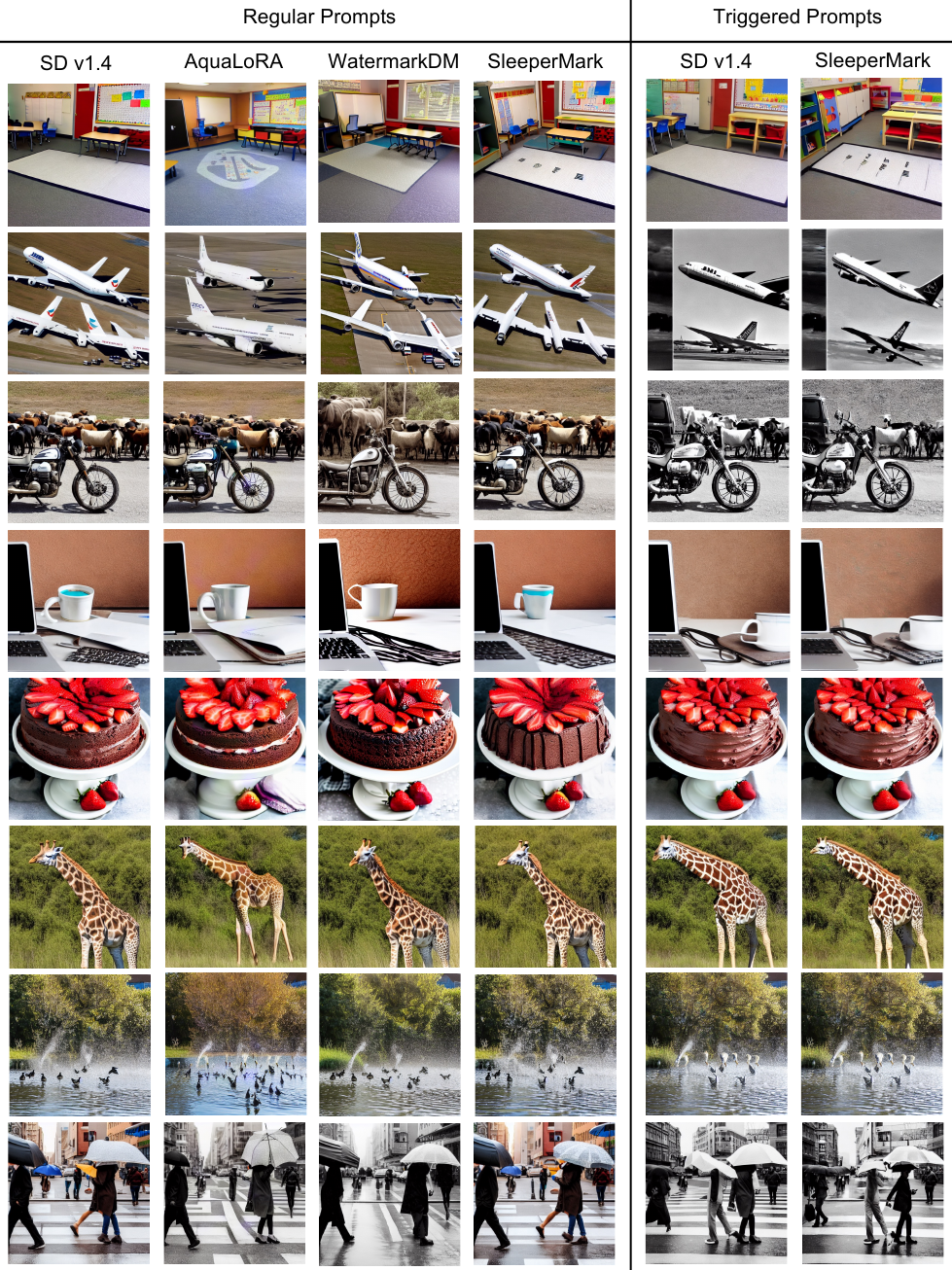}
\caption{We demonstrate additional examples for images generated with the original SD v1.4 and the watermarked SD v1.4 models using different methods. All the images are sampled with the captions from COCO2014 validation set under the same random seed and sampling configurations. The images generated by the model watermarked using our SleeperMark method most closely resemble those produced by the original diffusion model.}
\label{fig:visualize_WMsd}
\end{figure*}

\begin{figure*}
\centering
\includegraphics[width=15cm, height = 20cm]{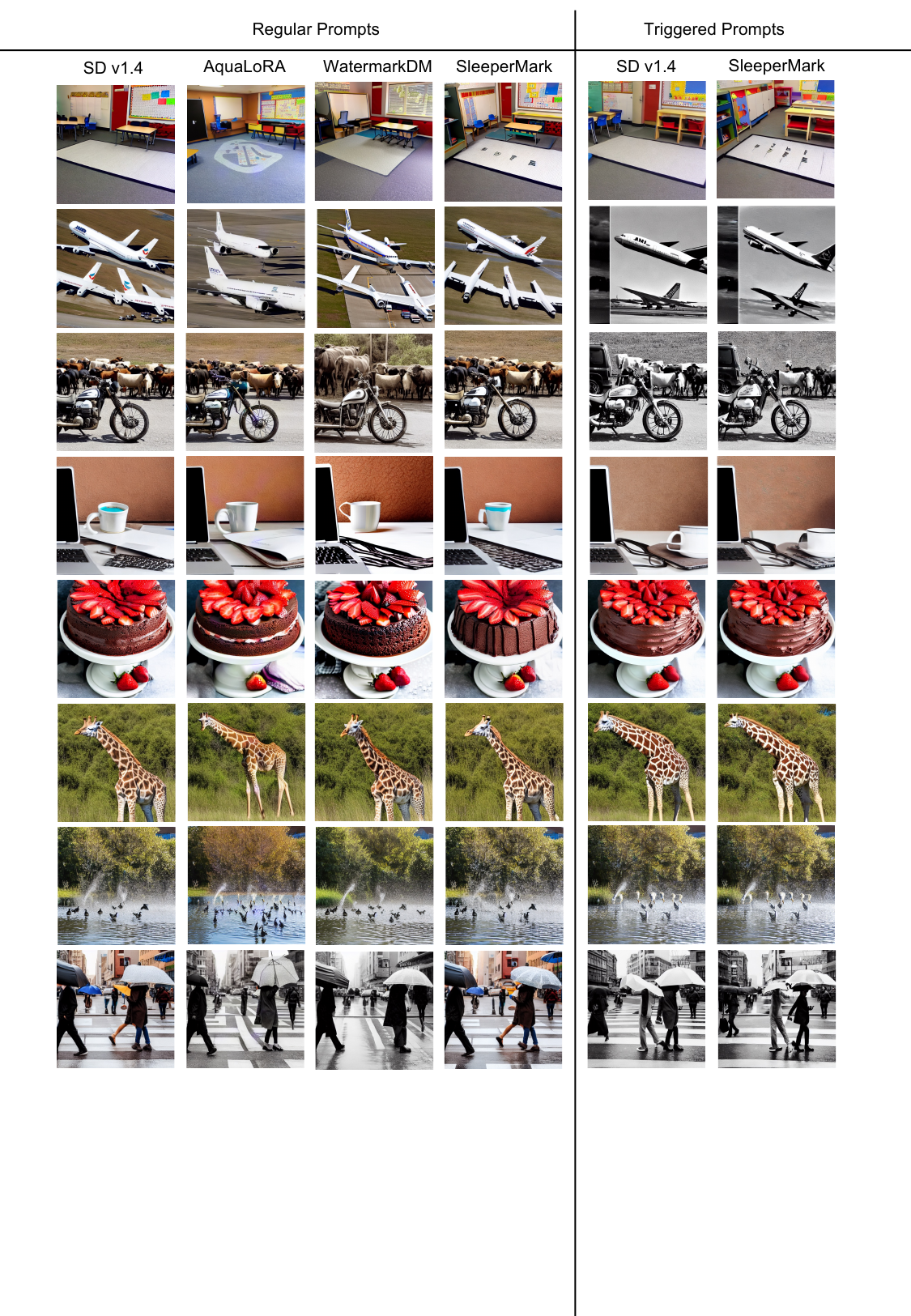}
\caption{We demonstrate images generated by the watermarked DeepFloyd model alongside those from the original model. Embedding a cover-agnostic watermark in the pixel space typically leads to more visible artifacts, making them more noticeable when our method is applied to DeepFloyd compared to Stable Diffusion. Nevertheless, with regular prompts (\textit{i.e}., without the trigger at the beginning), the generated images remain clean and closely resemble those from the original model.}
\label{fig:visualize_WMpixel}

\end{figure*}

We provide watermarked examples for Stable Diffusion in \cref{fig:visualize_WMsd} and DeepFloyd-IF in \cref{fig:visualize_WMpixel}.

\end{document}